\documentclass{article}

    \PassOptionsToPackage{numbers}{natbib}

    \usepackage[preprint]{neurips_2022}

\usepackage[utf8]{inputenc} 
\usepackage[T1]{fontenc}    
\usepackage{hyperref}       
\usepackage{url}            
\usepackage{booktabs}       
\usepackage{amsmath}
\usepackage{amsfonts}       
\usepackage{nicefrac}       
\usepackage{microtype}      
\usepackage{xcolor}         
\usepackage{multicol}
\usepackage{graphicx}
\usepackage{amsthm}
\usepackage{makecell}
\usepackage{array, multirow, graphicx}
\usepackage{float}

\newtheorem{theorem}{Theorem}[section]
\newtheorem{proposition}[theorem]{Proposition}
\newtheorem{definition}[theorem]{Definition}
\newtheorem{corollary}[theorem]{Corollary}

\newtheorem{remark}[theorem]{Remark}

\newtheorem{lemma}[theorem]{Lemma}
\title{PT$\Lp{p}$: Partial Transport $\Lp{p}$ Distances}

\author{%
Xinran Liu\thanks{Equal contribution}\\
  Department of Computer Science\\
  Vanderbilt University\\
  Nashville, TN 37235 \\
  \texttt{xinran.liu@vanderbilt.edu} \\
  \And 
  Yikun Bai\footnotemark[1]\\
  Department of Computer Science\\
  Vanderbilt University\\
  Nashville, TN 37235 \\
  \texttt{yikun.bai@vanderbilt.edu}
    \And 
  Huy Tran\\
  Department of Computer Science\\
  Vanderbilt University\\
  Nashville, TN 37235 \\
  \texttt{huy.tran@vanderbilt.edu}
  \And 
  Zhanqi Zhu\\
  Department of Computer Science\\
  Vanderbilt University\\
  Nashville, TN 37235 \\
  \texttt{zhanqi.zhu@vanderbilt.edu}
  \And 
  Matthew Thorpe\\
  Department of Mathematics\\
  The University of Manchester\\
  Manchester, UK \\
  \texttt{matthew.thorpe-2@manchester.ac.uk}
  \And 
  Soheil Kolouri\\
  Department of Computer Science\\
  Vanderbilt University\\
  Nashville, TN 37235 \\
  \texttt{soheil.kolouri@vanderbilt.edu}
}

\newcommand*\diff{\mathop{}\!\mathrm{d}}
\def\OPT{\mathrm{OPT}}
\def\SOPT{\mathrm{SOPT}}
\def\OT{\mathrm{OT}}
\def\SOT{\mathrm{SOT}}
\def\supp{\mathrm{supp}}
\def\Lp#1{\mathrm{L}^{#1}}
\def\TLp{\mathrm{TL}^p}
\def\TLpl{\mathrm{TL}^p_\beta}
\def\STLp{\mathrm{STL}^p}
\def\STLpl{\mathrm{STL}^p_\beta}
\def\PTLp#1{\mathrm{PTL}^{#1}}
\def\PTLpl{\mathrm{PTL}^p_{\beta,\lambda}}
\def\SPTLp{\mathrm{SPTL}^p}
\def\SPTLpl{\mathrm{SPTL}^p_{\beta,\lambda}}

\def\TV{\mathrm{TV}}

\def\dd{\mathrm{d}}
\def\id{\mathrm{id}}
\def\bbN{\mathbb{N}}
\def\bbR{\mathbb{R}}
\def\cM{\mathcal{M}}
\def\Ck#1{\mathrm{C}^{#1}}
\def\Ckb#1{\mathrm{C}_{\mathrm{b}}^{#1}}
\def\Ran{\mathrm{Ran}}
\def\lp{\left(}
\def\rp{\right)}
\def\lb{\left\{}
\def\rb{\right\}}
\def\spt{\mathrm{spt}}
\def\eps{\varepsilon}

\begin{document}

\maketitle

\begin{abstract}

Optimal transport and its related problems, including optimal partial transport, have proven to be valuable tools in machine learning for computing meaningful distances between probability or positive measures. This success has led to a growing interest in defining transport-based distances that allow for comparing signed measures and, more generally, multi-channeled signals. Transport $\Lp{p}$ distances are notable extensions of the optimal transport framework to signed and possibly multi-channeled signals. In this paper, we introduce partial transport $\Lp{p}$ distances as a new family of metrics for comparing generic signals, benefiting from the robustness of partial transport distances. We provide theoretical background such as the existence of optimal plans and the behavior of the distance in various limits. Furthermore, we introduce the sliced variation of these distances, which allows for rapid comparison of generic signals. Finally, we demonstrate the application of the proposed distances in signal class separability and nearest neighbor classification.

\end{abstract}

\section{Introduction}

At the heart of Machine Learning (ML) lies the ability to measure similarities or differences between signals existing in different domains, such as temporal, spatial, spatiotemporal grids, or even graphs in a broader sense. The effectiveness of any ML model depends significantly on the discriminatory power of the metrics it employs. Several criteria are desired when quantifying dissimilarities among diverse multivariate signals, including: 1) the ability to compare signals with varying lengths, 2) adherence to the inherent structure and geometry of the signals' domain, 3) being invariant to local deformation and symmetries, 4) computational efficiency, and 5) differentiability.  In recent literature, significant efforts have been dedicated to addressing these challenges. Prominent examples include the Dynamic Time Warping (DTW) \cite{sakoe1978dynamic} technique and its numerous extensions \cite{ddtw,ten2007multi,salvador2007toward,jeong2011weighted,cuturi2017soft}, as well as more recent methods based on optimal transport principles \cite{thorpe2017transportation,su2017order,janati2020spatio,zhang2020time}.

\textbf{Dynamic Time Warping  (DTW).} DTW is a technique for comparing and aligning time series signals that may vary in lengths or exhibit temporal distortions. To compare two signals, DTW computes the minimal-cost alignment between the signals \cite{sakoe1978dynamic}, enforcing the chronological order. The alignment problem in DTW is solved via dynamic programming (DP) using Bellman’s recursion, with quadratic cost in lengths of the signals. A large body of work has studied extensions of the DTW approach. For instance, Ten Holt et al. \cite{ten2007multi} extend DTW to multivariate time series. Salvador and Chan \cite{salvador2007toward} propose FastDTW, a linear time approximation of DTW with reasonable accuracy. To achieve robustness, Keogh and Pazzani \cite{ddtw} propose derivative DTW (DDTW), calculating the minimum-cost alignment based on derivatives of input signals, while Jeong et al. \cite{jeong2011weighted} consider the relative importance of alignments and propose weighted DTW (WDTW) providing robustness against outliers. Other notable extensions include Canonical Time Warping \cite{zhou2009canonical} and generalized time warping \cite{zhou2015generalized}, which enable the application of DTW to multi-modal sequences whose instances may have different dimensions. More recently, Cuturi \& Blondel \cite{cuturi2017soft} provide a differentiable variant of DTW, softDTW, allowing its seamless integration into end-to-end learning pipelines. 

\textbf{Optimal Transport.} Optimal transport (OT) has gained recognition as a powerful tool for quantifying dissimilarities between probability measures, finding broad applications in data science, statistics, machine learning, signal processing, and computer vision \cite{Villani2009Optimal,peyre2019computational}. The dissimilarity metrics derived from OT theory define a robust geometric framework for comparing probability measures, exhibiting desirable properties such as a weak Riemannian structure \cite{Villani2003Topics}, the concept of barycenters \cite{cuturi2014fast}, and parameterized geodesics \cite{ambrosio2005gradient}. However, it is important to note that OT has limitations when it comes to comparing general multi-channel signals. OT is specifically applicable to non-negative measures with equal total mass, restricting its use to signals that meet specific criteria: 1) single-channel representation, 2) non-negativity, and 3) integration to a common constant, such as unity for probability measures. In cases where signals do not fulfill these criteria, normalization or alternative methods are required for meaningful comparison using OT.


{\bf Unbalanced and Optimal Partial Transport.} Comparing non-negative measures with varying total amounts of mass is a common requirement in physical-world applications. In such scenarios, it is necessary to find partial correspondences or overlaps between two non-negative measures and compare them based on their respective corresponding and non-corresponding parts. Recent research has thus focused on extensions of the OT problem that enable the comparison of non-negative measures with unequal mass. The Hellinger-Kantorovich distance \cite{chizat2018interpolating, Liero2018Optimal}, optimal partial transport (OPT) problem \cite{caffarelli2010free,figalli2010optimal,figalli2010new}, Kantorovich-Rubinstein norm \cite{guittet2002extended,lellmann2014imaging} and unnormalized optimal transport~\cite{gangbo19,lee2021generalized} are some of the variants that fall under the category of "unbalanced optimal transport"  \cite{chizat2018interpolating, Liero2018Optimal}. These methods provide effective solutions for comparing non-negative measures in scenarios where the total amount of mass varies. It is important to note that although the unbalanced optimal transport methods have advanced the capabilities of comparing non-negative measures with unequal mass, they still cannot be used to compare multi-channel signals or signals with negative values. 

{\bf Transport-Based Comparison of Generic Signals.} Recent studies have proposed extensions of the Optimal Transport (OT) framework to compare multi-channel signals that may include negative values, while still harnessing the benefits of OT. For example, Su \& Hua \cite{su2017order} introduced the Order-preserving Wasserstein distance, which computes the OT problem between elements of sequences while ensuring temporal consistency through regularization of the transportation plan. A more rigorous treatment of the problem was proposed in \cite{thorpe2017transportation} that led to the so-called Transportation $\Lp{p}$ ($\TLp$) distances. In short, to compare two signals $f$ and $g$, $\TLp$ uses the OT distance between their corresponding measures, e.g., the Lebesgue measure, raised onto the graphs of the signals (See Section 3). Later, 
Zhang et al. \cite{zhang2020time} utilized a similar approach to $\TLp$ while adding entropy regularization \cite{cuturi2013sinkhorn} and introduced Time Adaptive OT (TAOT). Lastly, in Spatio-Temporal Alignments, Janati et al. \cite{janati2020spatio} combine OT with softDTW. They utilized regularized OT to capture spatial differences between time samples and employed softDTW for temporal alignment costs.

{\bf Contributions.} In this paper, we tackle the problem of comparing multi-channel signals using transport-based methods and present a new family of metrics, denoted as $\PTLp{p}$, based on the optimal partial transport framework. Our approach is motivated by the realization that while $\TLp$ distances allow for the comparison of general signals, they require complete correspondences between input signals, which limits their applicability to real-world signals that often exhibit partial correspondences. Our specific contributions are: 1) introducing a new family of metrics based on optimal partial transport for comparing multi-channel signals, 2) providing theoretical results on existence of the partial transport plan in the proposed metric, as well as the behavior of the distance in various limits, 3) providing the sliced variation of the proposed metric with significant computational benefits, and 4) demonstrating the robust performance of the proposed metric on nearest neighbor classification in comparison with various recent baselines. 

{\bf General Notations.}
We provide an extensive list of our notations in the supplementary material. Here we provide a small subset used in the development of our proposed framework. We use $\mathbb{R}_+$ for the set of postive real numbers, $\mathbb{R}^d$ to denote the d-dimensional Euclidean space, and $\mathbb{S}^{d-1}\subset \mathbb{R}^d$ to denote the unit hyper-sphere. Given $\Omega\subseteq\mathbb{R}^d,p\ge 1$, we use $\mathcal{P}(\Omega)$ to denote the set of Borel probability measures and  $\mathcal{P}_p(\Omega)$ to denote the set of probability measures with finite $p$'th moment defined on a metric space $(\Omega,d)$. We use $\mathcal{M}_+(\Omega)$ to denote the set of all positive Radon measures defined on $\Omega$. For $\mu\in \mathcal{P}_p(\Omega)$, we define $\Lp{p}(\mu;\mathbb{R}^k):=\{f:\Omega\rightarrow\mathbb{R}^k ~|~ \int_\Omega \|f(x)\|^p\diff\mu(x)<\infty\}$ to denote a Banach space with the usual norm. For $f:\Omega\rightarrow\hat{\Omega}$ and measure $\mu$ in $\mathcal{M}_+(\Omega)$ we use $f_\#\mu$ to denote the pushforward of measure $\mu$ through $f$, which is formally defined as $f_\#\mu(A)=\mu(f^{-1}(A))$ for $\forall A\subseteq \hat{\Omega}$. 

\section{Background - Optimal (Partial) Transport and Their Sliced Variations}
\noindent\textbf{Optimal Transport}. The OT problem in the Kantorovich formulation \cite{kantorovich1948problem} is defined for two probability measures $\mu$ and $\nu$ in $\mathcal{P}(\Omega)$, and a lower semi-continuous cost function $c:\Omega^2 \to \mathbb{R}+$ by:
\begin{equation}
\OT_c(\mu,\nu):=\inf_{\gamma\in\Pi(\mu,\nu)}\int_{\Omega^2} c(x,y)\diff\gamma(x,y),
\label{eq: OT_dist}
\end{equation}
Here, $\Pi(\mu,\nu)$ is the set of all joint probability measures whose marginals are $\mu$ and $\nu$. We represent this by $\pi_{1\#}\gamma=\mu$ and $\pi_{2\#}\gamma=\nu$, where $\pi_1$ and $\pi_2$ are the canonical projection maps.
 If $c(x,y)$ is a $p$-th power of a metric, then the $p$-th root of the resulting optimal value is known as the p-Wasserstein distance. This distance is a metric in $\mathcal{P}_p(\Omega)$. We will ignore the subscript $c$ if it is the default cost $\|\cdot\|^p$. Please see the appendix for more details.

\noindent\textbf{Optimal Partial Transport.} The problem of Optimal Partial Transport (OPT) extends the concept of mass transportation to include mass destruction at the source and mass creation at the target, with corresponding penalties for such actions. More precisely, let $\mu,\nu\in\mathcal{M}_+(\Omega)$, where $\mathcal{M}_+(\Omega)$ is set of positive Radon measures defined on $\Omega$. Let $\lambda\ge 0$ denote the penalty for mass creation or destruction. Then the OPT problem is defined as:
\begin{align}
\OPT_{\lambda,c}(\mu,\nu):=\inf_{\gamma\in\Pi_\leq(\mu,\nu)} \int_{\Omega^2} c(x,y)\diff\gamma(x,y)  +\lambda(\|\mu\|_{\TV}+\|\nu\|_{\TV}-2\|\gamma\|_{\TV})\label{eq: OPT_org}  
\end{align}
where $$ \Pi_\leq(\mu,\nu):=\{\gamma\in\mathcal{M}_+(\Omega^2): \pi_{1\#}\gamma\leq \mu, \pi_{2\#}\gamma\leq \nu\},
$$
$\pi_{1\#}\gamma \leq \mu$ indicates that $\pi_{1\#}\gamma$ is \textit{dominated by} $\mu$, i.e., for any Borel set $A\subseteq\Omega$, $\pi_{1\#}\gamma(A)\leq \mu(A)$, analogously for $\pi_{2\#}\gamma\leq \nu$. The cost function $c:\Omega^2\to \mathbb{R}$ is lower semi-continuous (generally, it is nonnegative), and $\|\mu\|_{\TV}$ is the total variation (and the total mass) of $\mu$, analogously for $\|\nu\|_{\TV},\|\gamma\|_{\TV}$.   
When the transportation cost $c(x,y)$ is a metric, $\OPT_{\lambda,c}(\cdot,\cdot)$ defines a metric on $\mathcal{M}_+(\Omega)$ (see~\cite[Proposition 2.10]{chizat2018unbalanced}, \cite[Proposition 5]{Piccoli2014Generalized}, \cite[Section 2.1]{lee2021generalized} and~\cite[Theorem 4]{chen2017matricial}). For simplicity of notation, we drop the $c$ in the subscript of $\OT$ and $\OPT$. 

\noindent\textbf{Sliced Transport.} For one-dimensional measures, i.e., when $\Omega\subseteq\mathbb{R}$, both OT and OPT problems have efficient solvers. In particular, the OT problem has a closed-form solution, and for discrete measures with $M$ and $N\geq M$ particles, it can be solved in $\mathcal{O}(N\log(N))$. Moreover, a quadratic algorithm, $\mathcal{O}(MN)$, was recently proposed in \citep{bai2022sliced} for the one-dimensional OPT problem. To extend the computational benefits of one-dimensional OT and OPT problems to d-dimensional measures, recent works utilize the idea of slicing, which is rooted in the Cramér–Wold theorem \cite{cramer1936some} and the Radon Transform from the integral geometry \cite{helgason2011integral,debnath2016integral}.  For $\theta\in\mathbb{S}^{d-1}$, a one-dimensional slice of measure $\mu\in\mathcal{M}_+(\Omega)$ can be obtained via $\langle\theta,\cdot\rangle_\#\mu$ where $\langle\cdot,\cdot\rangle:\Omega^2\to\mathbb{R}$ denotes the inner product. Then for $\mu,\nu\in\mathcal{P}_p(\Omega)$ we can define the Sliced-OT (SOT) as:
\begin{align}
    \SOT(\mu,\nu):=\int_{\mathbb{S}^{d-1}}\OT(\langle\theta,\cdot\rangle_\#\mu,\langle\theta,\cdot\rangle_\#\nu)\diff\sigma(\theta),
\end{align}
where $\sigma\in\mathcal{P}(\mathbb{S}^{d-1})$ is a probability measure such that $\supp(\sigma)=\mathbb{S}^{d-1}$, e.g., the uniform distribution on the unit hyper-sphere. Similarly, for $\mu,\nu\in\mathcal{M}_+(\Omega)$, Sliced-OPT (SOPT) \cite{bai2022sliced} can be defined as: 
\begin{align}
\SOPT_\lambda(\mu,\nu):=\int_{\mathbb{S}^{d-1}}\OPT_{\lambda(\theta)}(\langle\theta,\cdot\rangle_\#\mu,\langle\theta,\cdot\rangle_\#\nu)\diff\sigma(\theta),
\end{align}
where $\lambda\in \mathrm{L}^1(\sigma;\mathbb{R}_{+})$ is generally a projection dependent function.

\begin{figure}[t!]
    \centering    \includegraphics[width=\linewidth]{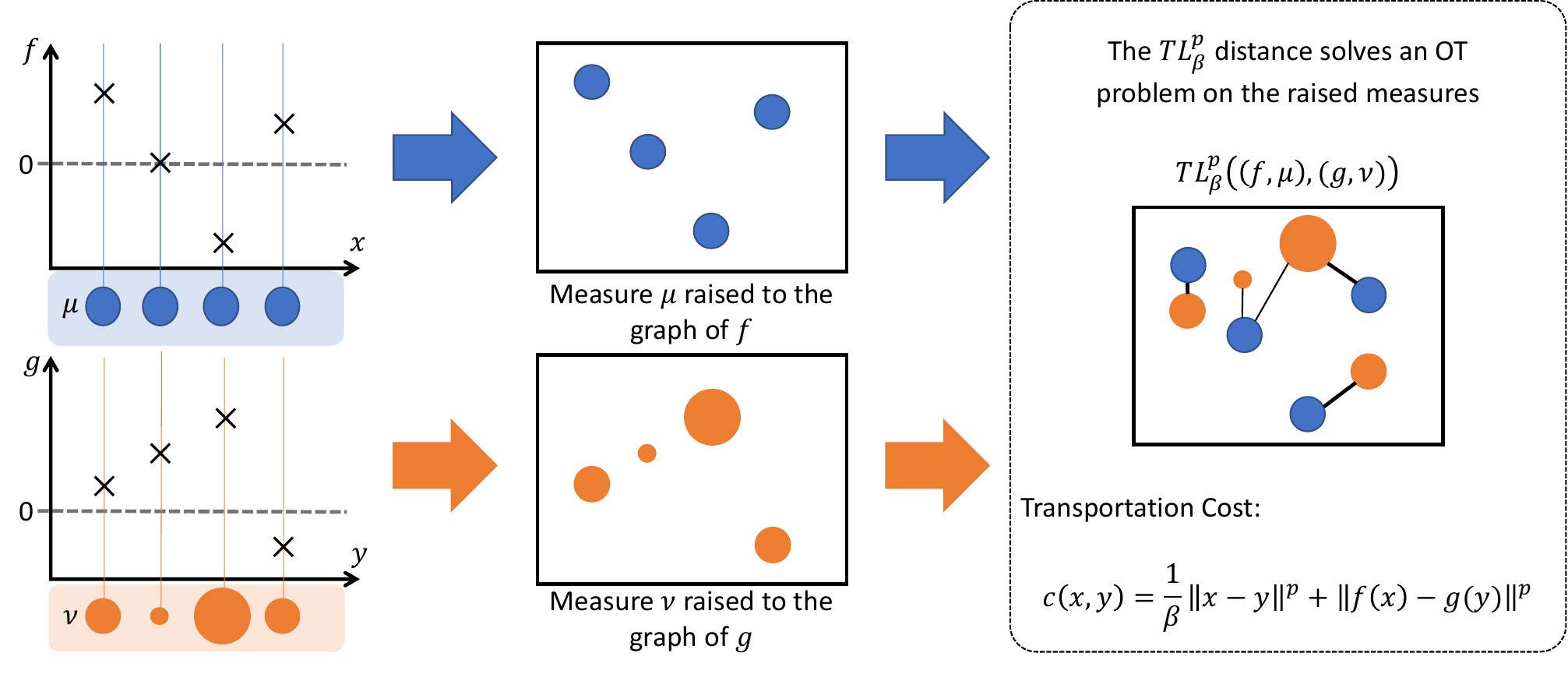}
    \caption{Illustrating the fundamental idea of $\TLp$ distances. On the left, signals $f$ and $g$ are depicted along with their associated measures $\mu$ and $\nu$. In the middle, the measures $\mu$ and $\nu$ are lifted to the graphs of $f$ and $g$, respectively. On the right, the optimal transport plan is visualized, accompanied by the corresponding transportation cost.}
    \label{fig:TLpEx}
\end{figure}

\section{Partial Transport for Multi-Channel Signals}

In the previous section, we discussed the suitability of OT and OPT problems (and similarly, SOT and SOPT problems) for comparing measures $\mu$ and $\nu$ in $\mathcal{P}_p(\Omega)$ or $\mathcal{M}_+(\Omega)$, respectively. In this section, we begin by defining a transport-based distance for multi-channel signals defined on a general class of measures, following the work of Thorpe et al. \cite{thorpe2017transportation} on Transport $\Lp{p}$ distances. We then motivate the need for partial transportation when comparing such multi-channel signals and introduce our Partial-Transport $\Lp{p}$, $\PTLp{p}$, distance.

\textbf{Transport $\Lp{p}$ Distances.} Following \cite{thorpe2017transportation}, a multi-channel signal with $k$ channels can be defined as the pair $(f,\mu)$ for $\mu\in \mathcal{P}_p(\Omega)$ and $f\in L^p(\mu;\mathbb{R}^k):=\{f:\Omega\to A\subseteq \mathbb{R}^k \}$. We denote the set of all such signals as 
$$\mathcal{Q}_p(\Omega;\mathbb{R}^k):=\{(f,\mu) | \mu\in \mathcal{P}_p(\Omega), f\in \Lp{p}(\mu;\mathbb{R}^k)\}.$$ We name it as the transport $\Lp{p}$ space. The $\TLpl$ distance between two such k-dimensional signals $(f,\mu)$ and $(g,\nu)$ in $\mathcal{Q}_p(\Omega;\mathbb{R}^k)$ is defined as: 
\begin{align}
    \TLpl((f,\mu),(g,\nu))=\inf_{\gamma\in\Pi(\mu,\nu)}\int_{\Omega^2}\big(\frac{1}{\beta}\|x-y\|^p+\|f(x)-g(y)\|^p\big)\diff \gamma(x,y)
    \label{eq:TLP}. 
\end{align}
 For any $p\in[1,\infty)$ and $\beta>0$, the $\TLpl$ distance defines a proper metric on $\mathcal{Q}_{p}(\Omega;\mathbb{R}^k)$, and $(\mathcal{Q}_{p}(\Omega;\mathbb{R}^k),\TLpl)$ is a metric space.  Intuitively, the $\TLpl$ measures the OT between measures $\mu$ and $\nu$ raised onto the graphs of $f$ and $g$. Hence, $\TLpl$ solves an OT problem in the $(d+k)$-dimensional space. Figure \ref{fig:TLpEx} shows the core concept behind $\TLp$ distances. Notably, the $\TLpl$ distance satisfies the following properties:
\begin{align}
\lim_{\beta\rightarrow 0} \TLpl((f,\mu),(g,\nu))&=\begin{cases}
    \|f-g\|_{\Lp{p}(\mu)}^p &\text{if }\mu=\nu \\ 
    \infty & \text{elsewhere}
\end{cases} \label{eq: TLP beta 0}\\
\lim_{\beta\rightarrow +\infty} \TLpl((f,\mu),(g,\nu))&=\OT(f_\#\mu,g_\#\nu)\label{eq: TLP beta infty} 
\end{align}
Hence, the $\TLpl$ distance interpolates between the $\Lp{p}$ distance between $f,g$ and the p-Wasserstein distance between $f_\# \mu$ and $g_\#\nu$.

\begin{figure}[t!]
    \centering
    \includegraphics[width=\linewidth]{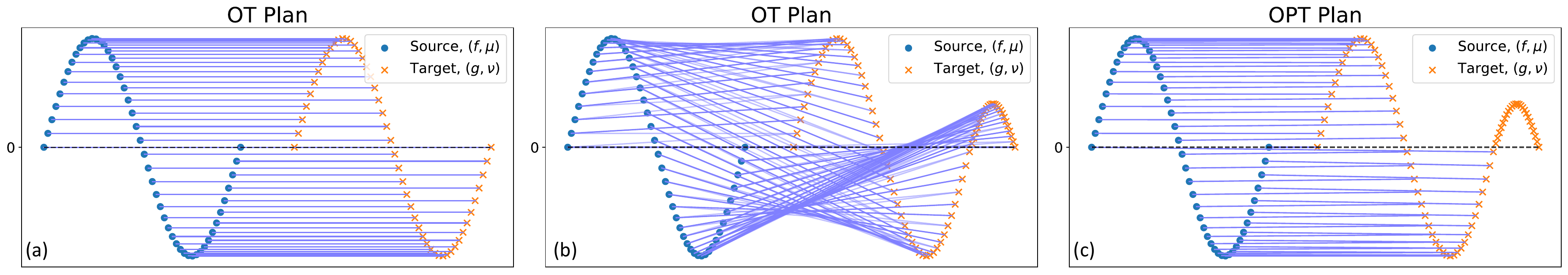}
    \caption{Highlighting the necessity of optimal partial transport. On the left, the signals exhibit complete correspondences, and OT successfully aligns the signals. In the middle, the signals display partial correspondences, and employing OT for comparison results in inaccurate correspondences. On the right, optimal partial transport (OPT) is employed, leading to the recovery of accurate partial correspondences.}
    \label{fig:ptlp}
\end{figure}

\textbf{Partial Transport $\Lp{p}$ Distances.} In many real-world scenarios, it is natural for two signals to only partially match each other. Figure \ref{fig:ptlp} illustrates this phenomenon. However, because $\TLp$ distances are rooted in OT, they may sacrifice true correspondences in order to achieve a complete match between the two signals (as seen in Figure \ref{fig:ptlp}). To address this issue, we propose extending the definition of $\TLp$ distances to partial transport, allowing for partial matching for signal comparison.

To do so, we first expand the definition of $k$-dimensional signals to be defined on positive measures rather than probability measures. Specifically, we define a signal as the pair $(f,\mu)$ where $\mu\in\mathcal{M}_+(\Omega)$ and $f\in \Lp{p}(f;\mathbb{R}^k)$. We denote the set of all such signals as $\mathcal{Q}_p^+(\Omega;\mathbb{R}^k)$,
that is,
$$\mathcal{Q}_p^+(\Omega;\mathbb{R}^k):=\{(f,\mu): \mu\in\mathcal{M}_+(\Omega), f\in \Lp{p}(\mu;\mathbb{R}^k)\}.$$

We now propose our Partial Transport $\Lp{p}$ ($\PTLp{p}$) distance between two signals $(f,\mu)$ and $(g,\nu)$ in $\mathcal{Q}_p^+(\Omega;\mathbb{R}^k)$ as:
\begin{align}
\PTLpl((f,\mu),(g,\nu))&=\inf_{\gamma\in\Pi_\leq(\mu,\nu)} \int_{\Omega^2} \left(\frac{1}{\beta}\|x-y\|^p+\|f(x)-g(y)\|^p\right)\diff\gamma(x,y) \nonumber\\ 
&\hspace{5em}+\lambda(\|\mu\|_{\TV}+\|\nu\|_{\TV}-2\|\gamma\|_{\TV})
    \label{eq: PTLP} 
    \end{align}

Note that as opposed to the $\TLp$ distance, the optimization in Eq. \eqref{eq: PTLP} is on the set of partial correspondences $\Pi_\leq(\mu,\nu)$. In practice, one often deals with discrete samples of $k$-dimensional signals. Consider $f$ and $g$ as two $k$-dimensional signals with $M$ and $N$ samples respectively. Let  $\mu=\sum_{i=1}^M\delta_{x_i}$ and $\nu=\sum_{j=1}^N \delta_{y_j}$ denote the empirical distributions, where $N, M\in\mathbb{N}$, $f_i=f(x_i)$, and $g_j=g(y_j)$. In this case, the $\PTLp{p}$ problem \eqref{eq: PTLP} can be written as: 
\begin{align}
\PTLp{p}_{\beta,\lambda}((f,\mu),(g,\nu))=&\sum_{\gamma\in\Pi_{\leq}(1_M,1_N)}(\frac{1}{\beta}\|x_i-y_j\|^p+\|f_i-g_j\|^p)\gamma_{ij} +\lambda(N+M-2|\gamma|) \label{eq: PTLP empirical}
\end{align}
where $1_M$ is the $M-$length vector with all 1 entries and analogously $1_N$; $$\Pi_\leq(1_M,1_N):=\{\gamma\in\mathbb{R}_+^{M\times N}:\gamma1_M\leq 1_N, \gamma^T 1_N\leq 1_M\};$$
and $|\gamma|=\sum_{ij}\gamma_{ij}$. In this case, we can further restrict the searching space of $\gamma$ as the optimal $\gamma$ will be induced by a 1-1 mapping.

Next, we provide some of the theoretical characteristics of $\PTLp{p}$. First, the PTLP problem \eqref{eq: PTLP} admits a minimizer, and the optimal value defines a metric in $\mathcal{Q}_p^+(\Omega;\mathbb{R}^k)$:  
\begin{theorem}\label{Thm: PTLP minimizer}
For any $p\geq 1$, and $\lambda,\beta>0$ there exists a minimizer for the $\PTLp{p}$ problem \eqref{eq: PTLP}. Furthermore, for the empirical $\PTLp{p}$ problem \eqref{eq: PTLP empirical}, there exists a minimizer $\gamma\in\Pi_{\leq}(1_M,1_N)$ that is induced by a 1-1 mapping. That is, the optimal $\gamma$ satisfies $\gamma_{ij}\in\{0,1\}$ for each $i,j$, and each row and column of $\gamma$ contains at most one nonzero element. 
\end{theorem}

\begin{theorem}\label{thm: PTLP is a metric}
$(\mathcal{Q}_+(\Omega;\mathbb{R}^k),\PTLpl)$ defines a metric space. 
\end{theorem}
We refer to Section \ref{sec: PTLP and OPT} in the appendix for the proofs of the above theorems and a detailed discussion of the $\PTLp{p}$ space $\mathcal{Q}_+(\Omega;\mathbb{R}^k)$. 

Similar to the $\TLp$ distance, we can also extend the definition for $\beta=0$ and $\beta=\infty$ by the following theorem: 
\begin{theorem}\label{thm: PTLP extreme beta}
If $\lambda>0$, we have 
\begin{align}
\lim_{\beta\to 0}\PTLp{p}_{\beta,\lambda}((f,\mu),(g,\nu))&=\|f-g\|_{\Lp{p}(\mu\wedge \nu),2\lambda}^p+\lambda(\|\mu-\nu\|_{\TV}) \label{eq: PTLP beta 0 0}\\
\lim_{\beta\to \infty}\PTLp{p}_{\beta,\lambda}((f,\mu),(g,\nu))&=\mathrm{OPT}_{\lambda}(f_\#\mu,g_\# \nu) \label{eq: PTLP beta infty 0},
\end{align}
where $\mu\wedge \nu$ is the minimum of measure $\mu,\nu$,
$$\|f-g\|_{\Lp{p}(\mu\wedge \nu),2\lambda}^p:=\int_{\Omega}\|f-g\|^p\wedge 2\lambda \, \dd(\mu\wedge \nu).$$
and $\|\mu-\nu\|_{\TV}$ is the total variation of the signed measure $\mu-\nu$. 
\end{theorem}
See Section \ref{sec: notations} in the appendix for the details of notations and Section \ref{sec: PTLP extreme beta} for the proof. Note, if we take $\lambda\to\infty$,  we can recover \eqref{eq: TLP beta 0}, \eqref{eq: TLP beta infty} by the above limits. We note that $\lambda\to 0$ is not an interesting case as it indicates zero cost for creation and destruction of mass, leading to an optimal $\gamma$ of all zeros, i.e., $\PTLp{p}_{\beta,0}((\mu,f),(\nu,g))=0$ for all $(\mu,f),(\nu,g)\in\mathcal{Q}^p_+(\Omega;\mathbb{R}^k)$. 


\textbf{Sliced Extensions of TLP and PTLP.}
Using the connection between the $\TLp$ distance and OT distance \cite{thorpe2017transportation}, Eq. \eqref{eq:TLP} can be rewritten as
\begin{align}
\TLp_{\beta}((f,\mu),(g,\nu))= \OT(\hat{\mu},\hat{\nu})
\label{eq: TLP-OT}
\end{align}
where $\hat{\mu}=(T_{\beta,f,p})_\# \mu$ is a push-forward measure of $\mu$ by $T_{\beta,f,p}(x)=\left[\begin{array}{c}x\beta^{-\frac{1}{p}}\\ f(x)\end{array}\right]$, and similarly $\hat{\nu}=(T_{\beta,g,p})_\# \nu$.
 Eq. \eqref{eq: TLP-OT} allows us to apply SOT method to the $\TLp$ distance, and have the sliced-TLP distance as follows: 
\begin{align}
\STLpl((f,\mu),(g,\nu))=\int_{\mathbb{S}^{d+k-1}} \OT(\theta_\#\hat{\mu},\theta_\#\hat{\nu})d\sigma(\theta)
\label{eq: sliced-TLP} 
\end{align}
where $\sigma(\theta)$ is a probability measure with non-zero density on $\mathbb{S}^{d+k-1}$, for instance the uniform measure on the unit sphere. 
Similarly, by leveraging SOPT and the relation between $\PTLp{p}$ and OPT (see proposition \ref{Pro: OPT and PTLP}), we can define Sliced $\PTLp{p}$ as
\begin{align}
\SPTLpl((f,\mu),(g,\nu))=\int_{\mathbb{S}^{d+k-1}} \mathrm{OPT}_{\lambda(\theta)}(\theta_\#\hat{\mu},\theta_\#\hat{\nu})d\sigma(\theta)
\label{eq:slicedPTLP}
\end{align}
where $\lambda$ can be defined as an $L^1(\sigma,\mathbb{R}_{++})$ function of $\theta$. Note that $\STLpl$ and $\SPTLpl$ are metrics on $\mathcal{Q}(\Omega;\mathbb{R}^k)$ and $\mathcal{Q}_+(\Omega;\mathbb{R}^k)$, respectively.

Equipped with the newly proposed distances, we now demonstrate their performance in separability and nearest neighbor classification. 
\section{Experiments}

\subsection{Separability}
A valid distance should be able to separate a mixture of different classes of signals. We aim to illustrate the separability of the $\PTLp{p}$ distance on different classes of signals in this experiment.\\

\textbf{Synthetic Data}\\
We generate the following two classes of signals on the domain $[0, 1]$:
\begin{align*}
&\mathbf{S}_0 = \{f(t)~|~f(t)=\varphi(t|x, \sigma_0);& \\
&\hspace{5em} x=0.98z+0.01, z\sim \mathrm{Unif}[0, 1]\}&\\
&\mathbf{S}_1 = \{g(t)~|~g(t)=\varphi(t|x+0.001, \sigma_1)-\varphi(t|x-0.001, \sigma_1);&\\
&\hspace{5em} x=0.98z+0.01, z\sim \mathrm{Unif}[0, 1]\}&\\
\end{align*}
where $\varphi$ denotes a Gaussian probability density function scaled within $[0, 1]$, $\sigma_0=0.01$ and $\sigma_1=\frac{0.01}{\sqrt{2}}$; time $t\sim \mathrm{Unif}[0, 1]$. In short, $\mathbf{S}_0$ is the class of signals with one positive Gaussian bump, whereas $\mathbf{S}_1$ denotes the class of signals with both a positive and a negative Gaussian bumps. To further test the robustness, we add random blip noise $\epsilon(t)$ to each signal as the second separability experiment:
\begin{align*}
    \epsilon(t) = \alpha\varphi(t|x, \sigma_e=0.001\sqrt{5}) + 0.1\epsilon_0
\end{align*}
where $\alpha$ is randomly chosen from $\{-0.5, 0.5\}$; $x=0.98z+0.01, z\sim \mathrm{Unif}[0, 1]$; $\epsilon_0$ is the Gaussian noise. $\epsilon(t)$ can be considered as a tiny positive/negative bump with Gaussian oscillation.\\
\begin{figure}[ht!]
    \centering
    \includegraphics[width=\textwidth]{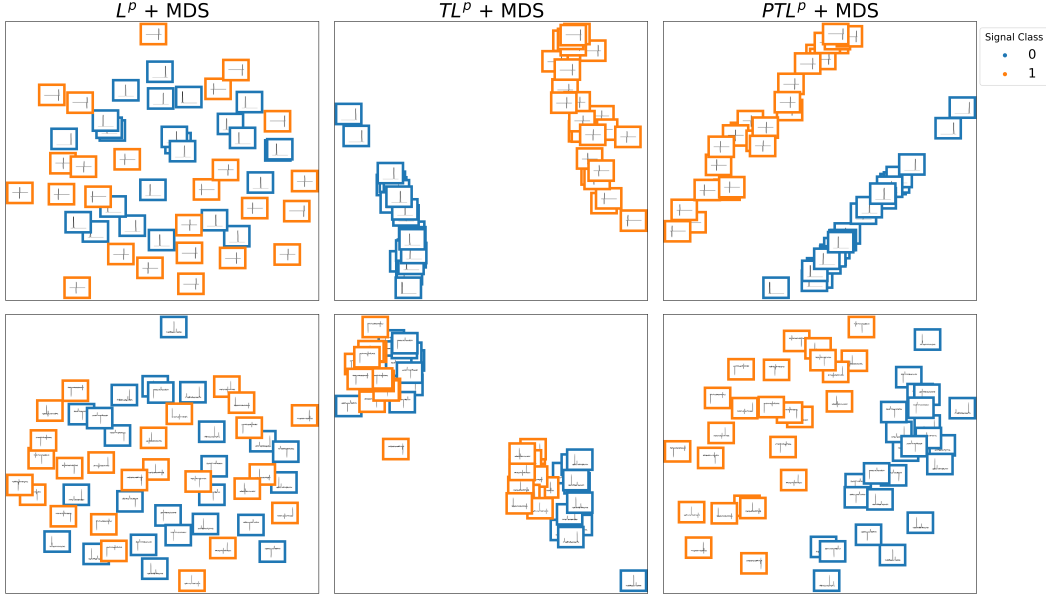}
    \caption{Visualizing manifold learning results for two classes of signals. For original signals (top row), both $\TLp$ and $\PTLp{p}$ separates two classes well, but $\Lp{p}$ fails. However, for the noisy signals (bottom row), only $\PTLp{p}$ shows a clear decision boundary.}
    \label{fig:synthetic}
\end{figure}

\textbf{Results}\\
Figure \ref{fig:synthetic} shows the 2D Multi-Dimensional Scaling (MDS) embeddings calculated from the precomputed pairwise $\Lp{p}$, $\TLp$ and $\PTLp{p}$ distance matrices. We observe that $\PTLp{p}$ not only achieves high performance in separating the two classes, but also exhibits robustness to noise. When adding blips, $\TLp$ tends to mistake the noise for the main trend and cluster signals based on the noise.

\subsection{1 Nearest Neighbor Classification}

\textbf{Experiment setup}\\
To demonstrate the effectiveness of our proposed $\PTLp{p}$ metric and its sliced variant $\SPTLp$, we test these methods on the task of 1 Nearest Neighbor (1NN) classification, along with other baselines. Given a test signal, we seek the nearest training signal with respect to each metric/divergence, and predict the test label as that of the found nearest neighbor. 

\textbf{Dataset}\\
We use three modified UCR datasets of varying lengths from \cite{tan2019time}: Suffix, Prefix and Subsequence. The Suffix dataset is generated by simulating scenarios when sensors are activated at different times, thus may miss some observations from the start and record only suffix time series. Similarly, the Prefix dataset generator imitates the sensor behavior of stopping non-deterministically and produces only prefix time series. The Subsequence dataset contains time series that have variations on both starting and stopping time, i.e. the sensor may only capture subsequences. 

\textbf{Baselines}\\
The $\Lp{p}$ distance between signals is known for its simplicity and efficiency, which fits signals in a fixed temporal grid.  OT-based similarity metrics, p-Wasserstein distance ($\OT$), and $\TLp$ treat signals / the graph of signals as probability measures and solve the optimization problem of transporting one probability measure to the other in the most cost-efficient way. Moreover, $\STLp$ is included in the baselines as a fast approximation of $\TLp$.

Unlike the $\Lp{p}$ metric, Dynamic Time Warping (DTW) \cite{sakoe1978dynamic} applies an elastic (non-linear) warping to temporal sequences and finds an optimal matching between the warped time series. DTW is more robust to time distortions by its pathological alignment. An $(N,M)$-warping path is a sequence $p=(p_1, p_2, \cdots, p_L)$ with $p_l=(n_l, m_l)\in [1:N]\times [1:M]$, which defines an alignment between two sequences of length $N$ and $M$ that satisfies monotonicity, continuity and boundary conditions \cite{Berndt1994UsingDT}. Given a pair of temporal sequences $f=\{f_i\}_{i=0}^N$ and $g=\{g_j\}_{j=0}^M$ on the domain $\Omega$, DTW is calculated as
\begin{equation}
    \label{eq:dtw}
    \mathrm{DTW}(f, g) = \min_p\{c_p(f, g)~|~p~\text{is an $(N,M)$-warping path}\},
\end{equation}
where $c_p(f, g)=\sum_{(i, j)\in p} c(f_i, g_j)$ and $c(f_i, g_j)$ is the cost of moving from $f_i$ to $g_j$. We also include variants of DTW, namely WDTW, DDTW, and Soft-DTW (SDTW) as baselines. For SDTW, we consider two cases for the smoothing parameter $\gamma=0.01$ and $\gamma=1$.

\textbf{Grid search for optimal $\beta$ and $\lambda$}\\
To find the optimal $\beta$ and $\lambda$ for $\PTLpl$, we perform grid search based on the 5-fold cross validation. We use the scikit-learn built-in GridSearchCV tools for implementation. The search range for $\beta$ is set to be $\{10^{-3}, 10^{-2}, 10^{-1}, 1, 10, 100, 10^3, 10^4\}$, and $\lambda$ is chosen from a set of 10 evenly spaced values from $0.1$ to the radius of the raised distribution on the graph of each signal.

In $\SPTLpl$, we also need to specify the slices, i.e. $\theta$'s for 1 dimensional projections. We obtain the optimal $\beta$ from  $\PTLpl$. As the amount of mass that should be transported may vary across slices, we adopt the strategy to search for the best $\lambda$ for the most informative slice, and then set $\lambda$'s accordingly for other slices. We set $\theta_0$ to be the first principle component of all signals. Note that $\theta_0$ vanishes at dimensions corresponding to $x\beta^{-\frac{1}{p}}$, but concentrates on $f(x)$ in $T_{\beta,f,p}(x)=\left[x\beta^{-\frac{1}{p}}; f(x)\right]$ (refer to Eq. \eqref{eq: TLP-OT} and Eq. \eqref{eq: sliced-TLP}). Similarly, we implement grid search for best $\lambda_{\theta_0}$ corresponding to $\theta_0$. Given $\theta_0$ and $\lambda_{\theta_0}$, for a specific slice $\theta$, $\lambda_{\theta} = \langle\theta, \theta_0\rangle\lambda_{\theta_0}$, where $\langle\cdot, \cdot\rangle$ denotes inner product.

\textbf{Results}\\
Table \ref{tab:tables} presents the results of nearest neighbor classification using different metrics/divergences on three subsets of the modified UCR dataset: Prefix, Subsequence, and Suffix. The table indicates that no single metric/divergence exhibits a significant advantage over others on a single dataset. However, $\SPTLp$ achieves the best performance on two out of three datasets and performs nearly as well as the top performers on the remaining dataset, resulting in an overall win. It is worth noting that although the improvement margins are small, the computational advantage of $\SPTLp$ and $\STLp$ compared to other competitors (see Figure \ref{fig:ptlp}), make them more favorable choices in terms of efficiency.

\begin{table}[!t]
    \label{tab:tables}
    \centering
    \caption{Nearest neighbor classification results on the modified UCR dataset \cite{tan2019time}. For each dataset the top two performers are turned bold. The average for each subset, i.e., Prefix, Subsequence, and Suffix, as well as the total average are reported. While the overall performances are close, we note that $\SPTLp$ and $\STLp$ provide significantly faster solutions when accelerated by parallel computation with respect to slices.}
    \small
    \scalebox{0.82}{
    \begin{tabular}{cc|c|c|c|c|c|c|c|c|c|c|c}
        \hline
        \multirow{18}{*}{{\rotatebox[origin=c]{90}{\textbf{Prefix Dataset}}}} & Method & $\PTLp{p}$& $\SPTLp$ & $\TLp$ & $\STLp$ & $\OT$&  \thead{SDTW\\{\tiny $\gamma=0.01$}}  &  \thead{SDTW\\{\tiny $\gamma=1$}} & DTW &WDTW & DDTW & $\Lp{p}$  \\ \hline
        & Adiac & 0.23 & \textbf{0.26} & 0.23 & 0.22 & 0.08 & 0.21 & 0.19 & 0.19 & 0.03 & \textbf{0.29} & 0.25 \\ 
        & ArrowHead & 0.49 & \textbf{0.51} & 0.44 & 0.41 & 0.45 & 0.45 & 0.45 & 0.43 & 0.39 & \textbf{0.52} & 0.25 \\ 
        & BeetleFly & 0.65 & \textbf{0.80} & 0.55 & 0.55 & 0.40 & 0.30 & 0.30 & 0.30 & 0.50 & \textbf{0.75} & 0.55 \\ 
        & DSR & \textbf{0.64} & 0.53 & 0.45 & 0.47 & 0.32 & 0.38 & 0.37 & 0.38 & 0.30 & 0.30 & \textbf{0.60} \\ 
        & DPOAG & \textbf{0.68} & \textbf{0.68} & 0.65 & 0.64 & 0.59 & 0.65 & 0.63 & 0.65 & 0.42 & 0.42 & 0.55 \\ 
        & ECG200 & \textbf{0.78} & 0.74 & \textbf{0.78} & 0.72 & 0.64 & 0.74 & 0.77 & 0.74 & 0.36 & 0.77 & 0.73 \\ 
        & Ham & 0.52 & \textbf{0.57} & 0.51 & 0.50 & 0.54 & 0.55 & 0.56 & \textbf{0.57} & 0.49 & 0.49 & 0.45 \\ 
        &    Herring & 0.63 & \textbf{0.69} & 0.61 & 0.61 & 0.47 & \textbf{0.67} & 0.66 & 0.66 & 0.59 & 0.59 & 0.56 \\ 
        & Lightning7 & 0.49 & 0.44 & 0.40 & 0.49 & 0.38 & \textbf{0.55} & 0.51 & \textbf{0.53} & 0.26 & 0.30 & 0.32 \\ 
        & OSULeaf & \textbf{0.45} & 0.41 & 0.40 & 0.33 & 0.32 & 0.34 & 0.42 & 0.34 & 0.10 & \textbf{0.52} & 0.19 \\ 
        & Plane & \textbf{0.84} & \textbf{0.80} & 0.69 & 0.68 & 0.48 & 0.68 & 0.72 & 0.67 & 0.14 & 0.14 & 0.50 \\ 
        & ShapeletSim & 0.52 & 0.49 & 0.48 & \textbf{0.53} & 0.50 & 0.51 & \textbf{0.57} & 0.51 & 0.50 & 0.50 & 0.49 \\ 
        & SyntheticControl & 0.84 & 0.81 & 0.76 & 0.83 & 0.55 & \textbf{0.87} & 0.84 & \textbf{0.87} & 0.17 & 0.40 & 0.48 \\ 
        & Trace & 0.75 & 0.72 & 0.79 & 0.75 & 0.44 & \textbf{0.81} & 0.78 & \textbf{0.82} & 0.23 & 0.24 & 0.61 \\ 
        & Wine & 0.50 & \textbf{0.59} & 0.46 & 0.54 & \textbf{0.59} & 0.44 & 0.46 & 0.46 & 0.50 & 0.56 & 0.50 \\ \hline
        &\thead{Average\\ Prefix}  & \textbf{0.59} & \textbf{0.60} & 0.55 & 0.55 & 0.45 & 0.54 & 0.55 & 0.54 & 0.33 & 0.45 & 0.47 \\ 
        \hline 
        \hline
        \multirow{15}{*}{{\rotatebox[origin=c]{90}{\textbf{Subsequence Dataset}}}}
        & Adiac & \textbf{0.17} & 0.14 & \textbf{0.17} & \textbf{0.17} & 0.05 & 0.12 & 0.12 & 0.12 & 0.03 & 0.18 & 0.05 \\ 
        & ArrowHead & 0.46 & \textbf{0.49} & \textbf{0.51} & 0.47 & 0.46 & 0.47 & 0.54 & 0.46 & 0.39 & 0.49 & 0.33 \\ 
       & BeetleFly & 0.55 & 0.50 & \textbf{0.65} & 0.55 & 0.60 & \textbf{0.65} & \textbf{0.65} & \textbf{0.65} & 0.50 & 0.50 & \textbf{0.65} \\ 
       & DSR & \textbf{0.39} & \textbf{0.39} & 0.34 & 0.34 & 0.34 & 0.43 & 0.38 & 0.35 & 0.30 & 0.31 & 0.29 \\ 
       & DPOAG & 0.63 & \textbf{0.65} & \textbf{0.65} & 0.63 & 0.54 & 0.60 & 0.58 & 0.60 & 0.42 & 0.60 & 0.54 \\ 
      &  ECG200 & \textbf{0.71} & \textbf{0.69} & 0.59 & 0.62 & 0.53 & 0.62 & 0.67 & 0.62 & 0.36 & 0.66 & 0.62 \\ 
      &  Ham & 0.48 & \textbf{0.54} & 0.49 & \textbf{0.52} & 0.42 & 0.46 & 0.49 & 0.46 & 0.49 & 0.49 & 0.44 \\ 
      &  Herring & 0.45 & 0.56 & 0.56 & \textbf{0.58} & 0.48 & 0.50 & 0.53 & 0.56 & \textbf{0.59} & 0.52 & 0.48 \\ 
      &  Lightning7 & 0.30 & 0.38 & 0.37 & \textbf{0.42} & 0.29 & 0.37 & \textbf{0.45} & 0.40 & 0.26 & 0.40 & 0.11 \\ 
     &   OSULeaf & \textbf{0.42} & 0.38 & 0.32 & 0.31 & 0.24 & 0.34 & 0.41 & 0.34 & 0.10 & \textbf{0.50} & 0.17 \\ 
       & Plane & \textbf{0.57} & 0.51 & 0.54 & \textbf{0.56} & 0.24 & 0.43 & 0.50 & 0.43 & 0.14 & 0.14 & 0.33 \\ 
      &  ShapeletSim & 0.49 & 0.46 & \textbf{0.53} & \textbf{0.53} & 0.51 & 0.53 & \textbf{0.58} & 0.53 & 0.50 & 0.52 & 0.47 \\ 
      &  SyntheticControl & 0.69 & 0.63 & 0.63 & 0.72 & 0.51 & \textbf{0.76} & \textbf{0.77} &\textbf{0.76} & 0.17 & 0.41 & 0.19 \\ 
      &  Trace & 0.63 & 0.56 & 0.67 & 0.68 & 0.36 & \textbf{0.78} & 0.64 & 0.74 & 0.23 & \textbf{0.82} & 0.44 \\ 
      &  Wine & 0.56 & \textbf{0.67} & \textbf{0.63} & 0.59 & 0.54 & 0.29 & 0.61 & 0.54 & 0.50 & 0.48 & 0.50 \\ \hline
      &  \thead{Average\\Subsequence} & 0.50 & 0.50 & 0.51 & \textbf{0.51} & 0.41 & 0.49 & \textbf{0.53} & 0.50 & 0.33 & 0.47 & 0.37 \\ 
      \hline
      \hline
      \multirow{15}{*}{{\rotatebox[origin=c]{90}{\textbf{Suffix Dataset}}}} & Adiac & 0.20 & 0.21 & \textbf{0.26} & \textbf{0.26} & 0.10 & 0.24 & 0.21 & 0.19 & 0.03 & \textbf{0.28} & 0.06 \\ 
      &  ArrowHead & 0.45 & \textbf{0.49} & 0.42 & 0.45 & 0.43 & 0.42 & 0.43 & 0.42 & 0.39 & \textbf{0.65} & 0.29 \\ 
      &  BeetleFly & 0.50 & 0.60 & 0.55 & \textbf{0.75} & 0.50 & 0.55 & \textbf{0.65} & 0.55 & 0.50 & 0.50 & 0.55 \\ 
      &  DSR & 0.51 & 0.57 & \textbf{0.62} & \textbf{0.62} & 0.37 & 0.43 & 0.43 & 0.43 & 0.30 & 0.23 & 0.38 \\ 
      &  DPOAG & 0.59 & 0.61 & \textbf{0.64} & 0.61 & 0.55 & \textbf{0.63} & 0.59 & 0.62 & 0.42 & 0.42 & 0.43 \\ 
      &  ECG200 & 0.68 & \textbf{0.75} & \textbf{0.72} & 0.70 & 0.65 & 0.70 & 0.71 & 0.69 & 0.36 & 0.67 & 0.67 \\ 
      &  Ham & 0.54 & \textbf{0.57} & 0.50 & 0.48 & 0.56 & 0.46 & \textbf{0.58} & 0.52 & 0.49 & 0.56 & 0.47 \\ 
      &  Herring & 0.56 & 0.56 & 0.52 & 0.52 & 0.47 & 0.56 & 0.50 & 0.48 & \textbf{0.59} & \textbf{0.59} & 0.58 \\ 
       & Lightning7 & 0.49 & 0.52 & 0.51 & \textbf{0.58} & 0.32 & 0.52 & \textbf{0.62} & 0.51 & 0.26 & 0.44 & 0.19 \\ 
      &  OSULeaf & 0.40 & 0.38 & 0.40 & 0.38 & 0.31 & 0.31 & \textbf{0.41} & 0.30 & 0.10 & \textbf{0.56} & 0.25 \\ 
     &   Plane & \textbf{0.70} & \textbf{0.70} & 0.64 & 0.63 & 0.48 & 0.67 & 0.68 & 0.67 & 0.14 & \textbf{0.75} & 0.20 \\ 
     &   ShapeletSim & 0.53 & \textbf{0.56} & \textbf{0.61} & 0.48 & \textbf{0.56} & 0.52 & 0.54 & 0.52 & 0.50 & 0.53 & 0.48 \\ 
     &   SyntheticControl & 0.77 & 0.76 & 0.76 & 0.78 & 0.51 & \textbf{0.86} & \textbf{0.87} & \textbf{0.86} & 0.17 & 0.44 & 0.10 \\ 
      &  Trace & 0.65 & 0.64 & 0.68 & 0.65 & 0.39 & \textbf{0.80} & 0.70 & \textbf{0.76} & 0.23 & 0.60 & 0.27 \\ 
     &   Wine & 0.56 & \textbf{0.57} & 0.56 & \textbf{0.57} & 0.56 & \textbf{0.57} & 0.54 & \textbf{0.59} & 0.50 & 0.54 & \textbf{0.57} \\ \hline
     &   \thead{Average\\ Suffix} & 0.54 & \textbf{0.57} & 0.56 & \textbf{0.56} & 0.45 & 0.55 & 0.56 & 0.54 & 0.33 & 0.52 & 0.37 \\ \hline
      & \thead{Average\\ Total} & \textbf{0.54} & \textbf{0.56} & 0.54 & 0.54 & 0.44 & 0.53 & 0.54 & 0.53 & 0.33 & 0.48 & 0.40 
    \\ \hline         
    \end{tabular}}
\end{table}

\subsection{Computation efficiency using Sliced $\PTLp{p}$}


\begin{table}[!t]
    \centering
    \caption{Worst case time complexities for our proposed methods and baselines. Here $N$ denotes the length of the signals, $d$ and $k$ are the signal dimension and number of channels respectively. $L$ is the number of slices for sliced methods. Note that DTW and its variants used in this paper share the same complexity, which is denoted by *DTW in the table.}
    \label{tab:complexity}
    \begin{tabular}{l|l}
    \hline
          Method & Worst-case Complexity\\
          \hline
          $\PTLp{p}$ &$\mathcal{O}(N^3(d+k))$\\
          $\SPTLp$ &$\mathcal{O}(LN((d+k)+N+\log(N)))$\\
          $\TLp$ & $\mathcal{O}(N^3(d+k))$\\
          $\STLp$ & $\mathcal{O}(LN((d+k)+\log(N)))$\\
          $\OT$  &  $\mathcal{O}(N^3k)$\\
          *DTW &  $\mathcal{O}(N^2k)$\\
          $\Lp{p}$ &  $\mathcal{O}(Nk)$\\
          \hline
    \end{tabular}
    
\end{table}

We summarize the time complexities of all methods considered in Table \ref{tab:complexity}. 

In implementation, DTW-based methods are solved by a dynamic programming algorithm. For DTW, soft-DTW, we use the solvers from \href{https://tslearn.readthedocs.io/en/stable/}{tslearn}, which are accelerated by \href{https://numba.pydata.org}{numba}. 
$\TLp$ and $\PTLp{p}$ are solved by linear programming solvers in \href{https://pythonot.github.io/}{PythonOT}, whose time complexity is cubic with respect to the length of signals in the worst case, and quadratic in practice when the measures are empirical. $\STLp$, $\SPTLp$ can be accelerated by \href{https://numba.pydata.org/https://numba.pydata.org/}{numba}. For $\STLp$ and $\SPTLp$, we set the number of projections to be 50. 
Note, the computation of $\STLp$ and $\SPTLp$ can be further accelerated by parallel computation with respect to slices. 

\newpage
\section{Conclusion}
In this paper, we propose partial transport $\Lp{p}$ ($\PTLp{p}$) distance as a similarity measure for generic signals. We have shown that $\PTLp{p}$ defines a metric that comes with an optimal transport plan. We further characterize the behaviors of $\PTLpl$ as $\beta$ goes to various limits. We extend $\PTLp{p}$ to sliced partial transport $\Lp{p}$ ($\SPTLp$), which is more computationally efficient. In the experimental section, we have demonstrated that the proposed metric is superior to other baselines in separability, and shown promising results on 1 nearest neighbor classification.

\newpage
\clearpage
\bibliography{optlp.bib}

\begin{thebibliography}{10}

\bibitem{sakoe1978dynamic}
Hiroaki Sakoe and Seibi Chiba.
\newblock Dynamic programming algorithm optimization for spoken word
  recognition.
\newblock {\em IEEE transactions on acoustics, speech, and signal processing},
  26(1):43--49, 1978.

\bibitem{ddtw}
Eamonn~J. Keogh and Michael~J. Pazzani.
\newblock {\em Derivative Dynamic Time Warping}, pages 1--11.

\bibitem{ten2007multi}
Gineke~A Ten~Holt, Marcel~JT Reinders, and Emile~A Hendriks.
\newblock Multi-dimensional dynamic time warping for gesture recognition.
\newblock In {\em Thirteenth annual conference of the Advanced School for
  Computing and Imaging}, volume 300, page~1, 2007.

\bibitem{salvador2007toward}
Stan Salvador and Philip Chan.
\newblock Toward accurate dynamic time warping in linear time and space.
\newblock {\em Intelligent Data Analysis}, 11(5):561--580, 2007.

\bibitem{jeong2011weighted}
Young-Seon Jeong, Myong~K Jeong, and Olufemi~A Omitaomu.
\newblock Weighted dynamic time warping for time series classification.
\newblock {\em Pattern recognition}, 44(9):2231--2240, 2011.

\bibitem{cuturi2017soft}
Marco Cuturi and Mathieu Blondel.
\newblock Soft-dtw: a differentiable loss function for time-series.
\newblock In {\em International conference on machine learning}, pages
  894--903. PMLR, 2017.

\bibitem{thorpe2017transportation}
Matthew Thorpe, Serim Park, Soheil Kolouri, Gustavo~K Rohde, and Dejan
  Slep{\v{c}}ev.
\newblock A transportation l\^{} p l p distance for signal analysis.
\newblock {\em Journal of mathematical imaging and vision}, 59:187--210, 2017.

\bibitem{su2017order}
Bing Su and Gang Hua.
\newblock Order-preserving wasserstein distance for sequence matching.
\newblock In {\em Proceedings of the IEEE conference on computer vision and
  pattern recognition}, pages 1049--1057, 2017.

\bibitem{janati2020spatio}
Hicham Janati, Marco Cuturi, and Alexandre Gramfort.
\newblock Spatio-temporal alignments: Optimal transport through space and time.
\newblock In {\em International Conference on Artificial Intelligence and
  Statistics}, pages 1695--1704. PMLR, 2020.

\bibitem{zhang2020time}
Zheng Zhang, Ping Tang, and Thomas Corpetti.
\newblock Time adaptive optimal transport: A framework of time series
  similarity measure.
\newblock {\em IEEE Access}, 8:149764--149774, 2020.

\bibitem{zhou2009canonical}
Feng Zhou and Fernando Torre.
\newblock Canonical time warping for alignment of human behavior.
\newblock {\em Advances in neural information processing systems}, 22, 2009.

\bibitem{zhou2015generalized}
Feng Zhou and Fernando De~la Torre.
\newblock Generalized canonical time warping.
\newblock {\em IEEE transactions on pattern analysis and machine intelligence},
  38(2):279--294, 2015.

\bibitem{Villani2009Optimal}
Cedric Villani.
\newblock {\em Optimal transport: old and new}.
\newblock Springer, 2009.

\bibitem{peyre2019computational}
Gabriel Peyr{\'e}, Marco Cuturi, et~al.
\newblock Computational optimal transport: With applications to data science.
\newblock {\em Foundations and Trends{\textregistered} in Machine Learning},
  11(5-6):355--607, 2019.

\bibitem{Villani2003Topics}
Cedric Villani.
\newblock {\em Topics in Optimal Transportation}.
\newblock American Mathematical Society, 2003.

\bibitem{cuturi2014fast}
Marco Cuturi and Arnaud Doucet.
\newblock Fast computation of wasserstein barycenters.
\newblock In {\em International conference on machine learning}, pages
  685--693. PMLR, 2014.

\bibitem{ambrosio2005gradient}
Luigi Ambrosio, Nicola Gigli, and Giuseppe Savar{\'e}.
\newblock {\em Gradient flows: in metric spaces and in the space of probability
  measures}.
\newblock Springer Science \& Business Media, 2005.

\bibitem{chizat2018interpolating}
Lenaic Chizat, Gabriel Peyr{\'e}, Bernhard Schmitzer, and Fran{\c{c}}ois-Xavier
  Vialard.
\newblock An interpolating distance between optimal transport and {Fisher--Rao}
  metrics.
\newblock {\em Foundations of Computational Mathematics}, 18(1):1--44, 2018.

\bibitem{Liero2018Optimal}
Matthias Liero, Alexander Mielke, and Giuseppe Savare.
\newblock Optimal entropy-transport problems and a new {Hellinger--Kantorovich}
  distance between positive measures.
\newblock {\em Inventiones mathematicae}, 211(3):969--1117, 2018.

\bibitem{caffarelli2010free}
Luis~A Caffarelli and Robert~J McCann.
\newblock Free boundaries in optimal transport and monge-ampere obstacle
  problems.
\newblock {\em Annals of mathematics}, pages 673--730, 2010.

\bibitem{figalli2010optimal}
Alessio Figalli.
\newblock The optimal partial transport problem.
\newblock {\em Archive for rational mechanics and analysis}, 195(2):533--560,
  2010.

\bibitem{figalli2010new}
Alessio Figalli and Nicola Gigli.
\newblock A new transportation distance between non-negative measures, with
  applications to gradients flows with dirichlet boundary conditions.
\newblock {\em Journal de math{\'e}matiques pures et appliqu{\'e}es},
  94(2):107--130, 2010.

\bibitem{guittet2002extended}
Kevin Guittet.
\newblock {\em Extended Kantorovich norms: a tool for optimization}.
\newblock PhD thesis, INRIA, 2002.

\bibitem{lellmann2014imaging}
Jan Lellmann, Dirk~A Lorenz, Carola Schonlieb, and Tuomo Valkonen.
\newblock Imaging with kantorovich--rubinstein discrepancy.
\newblock {\em SIAM Journal on Imaging Sciences}, 7(4):2833--2859, 2014.

\bibitem{gangbo19}
Wilfrid Gangbo, Wuchen Li, Stanley Osher, and Michael Puthawala.
\newblock Unnormalized optimal transport.
\newblock {\em Journal of Computational Physics}, 399:108940, 2019.

\bibitem{lee2021generalized}
Wonjun Lee, Rongjie Lai, Wuchen Li, and Stanley Osher.
\newblock Generalized unnormalized optimal transport and its fast algorithms.
\newblock {\em Journal of Computational Physics}, 436:110041, 2021.

\bibitem{cuturi2013sinkhorn}
Marco Cuturi.
\newblock Sinkhorn distances: Lightspeed computation of optimal transport.
\newblock {\em Advances in neural information processing systems}, 26, 2013.

\bibitem{kantorovich1948problem}
Leonid~V Kantorovich.
\newblock On a problem of monge.
\newblock In {\em CR (Doklady) Acad. Sci. URSS (NS)}, volume~3, pages 225--226,
  1948.

\bibitem{chizat2018unbalanced}
Lenaic Chizat, Gabriel Peyr{\'e}, Bernhard Schmitzer, and Fran{\c{c}}ois-Xavier
  Vialard.
\newblock Unbalanced optimal transport: Dynamic and {Kantorovich} formulations.
\newblock {\em Journal of Functional Analysis}, 274(11):3090--3123, 2018.

\bibitem{Piccoli2014Generalized}
Benedetto Piccoli and Francesco Rossi.
\newblock Generalized wasserstein distance and its application to transport
  equations with source.
\newblock {\em Archive for Rational Mechanics and Analysis}, 211(1):335--358,
  2014.

\bibitem{chen2017matricial}
Yongxin Chen, Tryphon~T Georgiou, Lipeng Ning, and Allen Tannenbaum.
\newblock Matricial wasserstein-1 distance.
\newblock {\em IEEE control systems letters}, 1(1):14--19, 2017.

\bibitem{bai2022sliced}
Yikun Bai, Bernard Schmitzer, Mathew Thorpe, and Soheil Kolouri.
\newblock Sliced optimal partial transport.
\newblock {\em arXiv preprint arXiv:2212.08049}, 2022.

\bibitem{cramer1936some}
Harald Cram{\'e}r and Herman Wold.
\newblock Some theorems on distribution functions.
\newblock {\em Journal of the London Mathematical Society}, 1(4):290--294,
  1936.

\bibitem{helgason2011integral}
Sigurdur Helgason et~al.
\newblock {\em Integral geometry and Radon transforms}.
\newblock Springer, 2011.

\bibitem{debnath2016integral}
Lokenath Debnath and Dambaru Bhatta.
\newblock {\em Integral transforms and their applications}.
\newblock Chapman and Hall/CRC, 2016.

\bibitem{tan2019time}
Chang~Wei Tan, Fran{\c{c}}ois Petitjean, Eamonn Keogh, and Geoffrey~I Webb.
\newblock Time series classification for varying length series.
\newblock {\em arXiv preprint arXiv:1910.04341}, 2019.

\bibitem{Berndt1994UsingDT}
Donald~J. Berndt and James Clifford.
\newblock Using dynamic time warping to find patterns in time series.
\newblock In {\em KDD Workshop}, 1994.

\bibitem{garcia2016continuum}
Nicol{\'a}s Garc{\'\i}a~Trillos and Dejan Slep{\v{c}}ev.
\newblock Continuum limit of total variation on point clouds.
\newblock {\em Archive for rational mechanics and analysis}, 220:193--241,
  2016.

\bibitem{heinemann2023kantorovich}
Florian Heinemann, Marcel Klatt, and Axel Munk.
\newblock Kantorovich--rubinstein distance and barycenter for finitely
  supported measures: Foundations and algorithms.
\newblock {\em Applied Mathematics \& Optimization}, 87(1):4, 2023.

\end{thebibliography}
\bibliographystyle{unsrt}
\newpage 

\newpage\clearpage

\section{Appendix}

\appendix 
We refer to the main text for the references. 
\section{Notation}
\label{sec: notations}
\begin{itemize}
\item $\mathbb{R}_+:=\{x\in \mathbb{R}: x\ge 0\}$.
\item $\mathbb{R}_{++}:=\{x\in\mathbb{R}: x>0\}$.
\item $\mathbb{R}^k$: the codomain of signals, where $k\ge 1$.  
\item $\Omega$: unless otherwise stated is a closed subset of $\mathbb{R}^d$ where $d\ge 1$.
\item $\mathcal{M}(\Omega)$, $\mathcal{M}(\Omega^2)$: the set of signed Radon measures on $\Omega$, $\Omega^2$ respectively. 
\item $\mathcal{M}_+(\Omega)$, $\mathcal{M}_+(\Omega^2)$: the set of positive Radon measures on $\Omega$, $\Omega^2$ respectively.  
\item $\spt(\mu)$ for $\mu\in\cM(\Omega)$: the support of the measure $\mu$.
\item $\|\mu\|_{\TV}$ where $\mu\in\mathcal{M}(\Omega)$: the total variation of $\mu$. If $\mu$ is positive, $\|\mu\|_{\TV}=\mu(\Omega)$. 
\item Weak-convergence of measures:
given a sequence $\{\gamma_n\}_{n\in\bbN}\subset \mathcal{M}(\Omega^2)$, if there exists $\gamma_0\in \mathcal{M}(\Omega^2)$ such that 
$$\int f \,\dd\gamma_n\to \int f \,\dd\gamma,\forall f\in \Ck{0}(\Omega^2),$$ 
then we say $\gamma_n$ converges weakly in measure to $\gamma$, and write 
$$\gamma_n{\rightharpoonup}\gamma.$$ 
Note that this type of convergence is also an example of weak$^*$ convergence.
\item Sequential compactness in the weak topology: given $K\subset \mathcal{M}_+(\Omega^2)$, $K$ is said to be sequentially compact in the weak topology if for any sequence $\{\gamma_n\}_{n\in\mathbb{N}}\subset K$, there exists a further subsequence and some $\gamma\in K$ such that $\gamma_n{\rightharpoonup} \gamma$. 
\item $\pi_1,\pi_2$: the canonical projections on $\Omega^2$, i.e. $\pi_1((x,y))=x,\pi_2((x,y))=y$.
\item $f_\# \gamma$: the push-forward of the measure $\gamma$ by $f$, i.e. $f_\#\gamma(A)=\gamma(f^{-1}(A))$ for any Borel set $A$. 
\item $\mu\wedge\nu$: minimum measure between $\mu$ and $\nu$. Formally, for any Borel set $A$, 
$$\mu\wedge\nu(A):=\inf\{\mu(A_1)+\nu(A_2)\},$$
where the infimum is taken over all possible partitions of $A=A_1\cup A_2$. For example, if $\mu,\nu$ are continuous with respect to some reference measure $\mathcal{L}$, say $\mu=f_\mu \mathcal{L}, \nu=f_\nu \mathcal{L}$, then $\mu\wedge\nu =(f_\mu \wedge f_\nu) \mathcal{L}$, where $f_\mu\wedge f_\nu$ denotes the minimum between $f_\mu(x)$ and $f_\nu(x)$ for each $x$.  
\item $\mu\leq \nu$ where $\mu,\nu\in\mathcal{M}_+(\Omega)$: if for each Borel set $A\subset\Omega$, $\mu(A)\leq \nu(A)$. 
\item $\Pi(\mu,\nu)$ where $\mu,\nu\in \mathcal{M}_+(\Omega)$: the set of Kantorovich plans between $\mu,\nu$, i.e. $\Pi(\mu,\nu):=\{\gamma\in \mathcal{M}_+(\Omega^2): (\pi_1)_\# \gamma=\mu,(\pi_2)_\#\gamma=\nu\}$. Note, the set is nonempty if and only if $\mu(\Omega)=\nu(\Omega)$. 
\item $\Pi_\leq(\mu,\nu)$ where $\mu,\nu\in \mathcal{M}_+(\Omega)$: the set of Kantorovich plans between $\mu,\nu$ for the $\OPT$ problem, i.e. $\Pi(\mu,\nu):=\{\gamma\in \mathcal{M}_+(\Omega^2): (\pi_1)_\# \gamma\leq\mu,(\pi_2)_\#\gamma\leq\nu\}$. 
\item $(\Ckb{0}(\Omega^2),\|\cdot\|_{\sup})$: the space of continuous and bounded functions on $\Omega^2$. 
Note, the dual space of $\Ckb{0}(\Omega^2)$ is the space of Radon measures $\mathcal{M}(\Omega^2)$. 
\item $\|\cdot\|_{\Lp{p}(\mu)},\|\cdot\|_{\Lp{p}(\mu),2\lambda}$ 
where $\mu\in \mathcal{M}_+(\Omega)$: the $\Lp{p}$ norm and truncated $\Lp{p}$ norm with respect to reference measure $\mu$.  
i.e.
\begin{align*}
\|f\|_{\Lp{p}(\mu)} & = \left( \int_\Omega \|f(x)\|^p \, \dd\mu(x)\right)^{1/p}, \\
\|f\|_{\Lp{p}(\mu),2\lambda} & = \left(\int_\Omega \|f(x)\|^p\wedge2\lambda \, \dd \mu(x)\right)^{1/p}
\end{align*}
\item $f,g :\Omega\to \mathbb{R}^k$: functions that represent the signals.  
\item $\beta>0,\lambda\ge 0$: constants used in the $\PTLp{p}$ problem.
\item $p\ge 1$: a constant which denotes the order of norm. 
\item $\mathcal{Q}_p^+(\Omega;\mathbb{R}^k)$: the $\PTLp{p}$ space, we refer to Section \ref{sec: PTLP and OPT} for the definition.
\item $D_{\beta}:(\Omega\times\mathbb{R}^k)^2\to \mathbb{R}$ with 
$$D^p_\beta((x,\tilde{x}),(y,\tilde{y}))=\frac{1}{\beta}\|x-y\|^p+\|\tilde{x}-\tilde{y}\|^p:$$
the cost function in the $\PTLp{p}$ problem.
Note, $D_{\beta}$ defines a metric and is equivalent to $\ell^p$ norm on $\Omega\times\mathbb{R}^k$.
\item $C(\gamma;(f,\mu),(g,\nu),\beta,\lambda)=\int_{\Omega^2} D_\beta((x,f(x)),(y,g(y))) \, \dd \gamma(x,y) + \lambda(\|\mu\|_{\TV}+\|\nu\|_{\TV}-2\|\gamma\|_{\TV})$: the objective function in the $\PTLp{p}$ problem \eqref{eq: PTLP}.
\item $C^\prime(\gamma;(f,\mu),(g,\nu),\beta,\lambda)=\int_{\Omega^2}D_\beta((x,f(x)),(y,g(y))) \wedge 2\lambda \, \dd\gamma(x,y) + \lambda(\|\mu\|_{\TV}+\|\nu\|_{\TV}-2\|\gamma\|_{\TV})$: the objective function in an equivalent formulation of the $\PTLp{p}$ problem as defined in \eqref{eq: PTLP 2}.
\item $\hat{C}(\hat\gamma; \hat\mu,\hat\nu,D_\beta^p,\lambda)=\int_{\Omega^2}D_\beta^p((x,\tilde{x}),(y,\tilde{y})))\,\dd\gamma+\lambda(\|\hat\mu\|_{\TV}+\|\hat\nu\|_{\TV}-2\|\hat\gamma\|_{\TV})$: the objective function in the OPT problem \eqref{eq: OPT 2}.
\item $c_{\beta,f,g}(x,y):=\frac{1}{\beta}\|x-y\|^p+\|f(x)-g(y)\|^p$, $c_{\infty,f,g}(x,y):=\|f(x)-g(y)\|^p$: the cost function in the OPT problems \eqref{eq: OPT 3}, \eqref{eq: OPT 4} 
\item $\hat\mu:=(\id\times f)_\# \mu,\hat\nu:=(\id\times g)_\# \nu,\hat\gamma:=((\id\times f),(\id\times g))_\#\gamma$, where $\mu,\nu\in\mathcal{M}_+(\Omega), \gamma\in \mathcal{M}_+(\Omega^2)$: the identification between the $\PTLp{p}$ space, $\mathcal{Q}_p^+(\Omega;\mathbb{R}^k)$, and a subset of $\mathcal{M}_+(\Omega\times \bbR^k)$. 
\item $\text{OT}_c(\mu,\nu),\text{OT}(\mu,\nu)$: the optimal transportation problem, where $c$ is the cost function. The default cost is the (p-th power of the) $\ell^p$ norm, and we use  $\text{OT}(\mu,\nu)$ to denote 
$\text{OT}_{\|\cdot\|^p}(\mu,\nu)$.
\item $\text{OPT}_c(\mu,\nu),\text{OPT}(\mu,\nu)$: the optimal partial transportation problem, where $c$ is the cost function. Analog to $\text{OT}_c,\text{OT}$. 
\end{itemize}

\section{Preliminary Results for Radon Measures}

We start by showing that weak convergence in measure preserves inequalities.

\begin{proposition}\label{cor: bounded sequence}
Given a sequence $(\mu_n)_{n\in\mathbb{N}}\subset\mathcal{M}_+(\Omega)$, if $\mu_n{\rightharpoonup}\mu_0$ for some $\mu_0\in \mathcal{M}(\Omega)$, then $\mu_0\in\mathcal{M}_+(\Omega)$ and $\mu_0\leq \mu$.
\end{proposition}

\begin{proof}
Pick any continuous and bounded function $f\in \Ckb{0}(\Omega)$, we have 
$$\int f \, \dd\mu_0=\lim_{n\to\infty}\int_\Omega f\, \dd\mu_n\leq \int_\Omega f \, \dd\mu,$$
thus $\mu_0\leq \mu$. 
Similarly, we have $\mu_0\ge 0$. 
\end{proof}

We will also make use of the closure, in the weak topology, of the space $\Pi_{\leq}(\mu,\nu)$, which follows similarly to the analogous result for the space $\Pi(\mu,\nu)$.
A similar result has been proved when $\mu,\nu$ are continuous measure,  for example \cite[Lemma 2.2]{caffarelli2010free}.
For completeness we include a proof in the general case.

\begin{proposition}\label{pro: compact set}
The set $\Pi_\leq(\mu,\nu)$ is sequentially compact in the weak topology. 
\end{proposition}

\begin{proof} 
Pick a sequence $\{\gamma_n\}_{n\in\bbN}\subset \Pi_\leq(\mu,\nu)\subset\mathcal{M}_+(\Omega^2)\subset\mathcal{M}(\Omega^2)$. 
We have $\gamma_n$ is bounded with respect to total variation.
Indeed, 
\[ \|\gamma_n\|_{\TV} = \gamma_n(\Omega^2) \leq \mu(\Omega), \quad \forall n \]
In addition, we will show $\gamma_n$ is tight.  
Pick $\epsilon>0$, since $\mu,\nu$ are inner regular, there exists compact set $K\subset \Omega$ such that $\mu(\Omega\setminus K),\nu(\Omega\setminus K)\leq \epsilon$. 
$$\gamma_n(\Omega\setminus K^2)\leq  \gamma_n(\Omega\times (\Omega\setminus K)+\gamma_n((\Omega\setminus K)\times\Omega)\leq \mu(\Omega\setminus K)+\nu(\Omega\setminus K)\leq 2\epsilon.$$
Thus $(\gamma_n)$ is a tight sequence. 

By Prokhorov's theorem for signed measures the closure (in the weak topology) of $\Pi_{\leq}(\mu,\nu)$ is weakly sequentially compact in $\cM(\Omega^2)$. 
It remains to show $\Pi_{\leq}(\mu,\nu)$ is weakly closed. 

Let $\gamma_n\rightharpoonup\gamma\in\cM(\Omega^2)$.
First, we claim 
$\pi_{1\#}\gamma_n \rightharpoonup \pi_{1\#} \gamma, \pi_{2\#}\gamma_n \rightharpoonup \pi_{2\#}\gamma$.
Indeed, pick $f\in \Ckb{0}(\Omega)$, $f$ can be regarded as a function in $\Ckb{0}(\Omega^2)$ whose value is independent to the second input $y$. Thus, we have 
$$ \lim_{n\to\infty} \int_{\Omega} f(x) \, \dd \pi_{1\#} \gamma_n(x) = \lim_{n\to\infty} \int_{\Omega^2} f(x) \, \dd \gamma_n(x,y) = \int_{\Omega^2} f(x) \, \dd \gamma(x,y) = \int_{\Omega} f(x) \, \dd \pi_{1\#} \gamma(x) $$
that is $\pi_{1\#}\gamma_n \rightharpoonup \pi_{1\#} \gamma$ and similarly $\pi_{2\#}\gamma_n \rightharpoonup \pi_{2\#} \gamma$. 
By Lemma \ref{cor: bounded sequence}, we have $\pi_{1\#}\gamma\leq \mu, \pi_{2\#}\gamma\leq \nu$ and $\gamma\ge 0$. 
\end{proof}


\section{Relation between \texorpdfstring{$\PTLp{p}$}{PTLp} and OPT} \label{sec: PTLP and OPT}

We formally introduce the $\PTLp{p}$ space.

\begin{definition}
Given nonempty closed $\Omega\subset\mathbb{R}^d$
the $\PTLp{p}$ space is defined as 
$$\mathcal{Q}_{p}^+(\Omega;\mathbb{R}^k)=\{(f,\mu): \mu\in \mathcal{M}_+(\Omega), f\in \Lp{p}(\mu;\mathbb{R}^k)\}.$$
The identity in $\mathcal{Q}_p^+(\Omega;\mathbb{R}^k)$ is defined as:
$(f,\mu)=(g,\nu)$ if and only if 
$\mu=\nu$ and $f=g, \mu$-a.s.  
\end{definition}

Inspired by the technique in \cite{garcia2016continuum}, it is easy to see that the mapping 
$T: \mathcal{Q}_p^+(\Omega;\mathbb{R}^k)\to \mathcal{M}_+(\Omega\times \mathbb{R}^k)$ with 
$(f,\mu)\mapsto (\id\times f)_\#\mu$ defines an embedding. That is, the map $T$ allows us to identify an element $(f,\mu)\in \mathcal{Q}_p^+(\Omega;\mathbb{R}^k)$ with a measure in the product space $\mathcal{M}_+(\Omega\times\mathbb{R}^k)$ which can be written as $(\id\times f)_\# \mu$. 

\begin{proposition}\label{pro: OPT and PTLP mapping}
The mapping $T: \mathcal{Q}_p^+(\Omega;\mathbb{R}^k) \to \Ran(T)\subset \mathcal{M}_+(\Omega\times \mathbb{R}^k)$ is a 1-1 mapping. 
\end{proposition}

\begin{proof}
Choose distinct $(f,\mu),(g,\nu)\in \mathcal{Q}^+_p(\Omega;\mathbb{R}^k)$. Then by the identity in $\mathcal{Q}^+_p(\Omega;\mathbb{R}^k)$ as we defined above, we have one of the following: 
$\mu\neq \nu$ or $\mu=\nu$ and $\mu(x: f(x)\neq g(x))>0$. 

For the first case, there exists a Borel set $A\subset \Omega$ such that $\mu(A)\neq \nu(A)$. Without loss of generality, we suppose $\mu(A)>\nu(A)$. 
We have 
\begin{align}
T((f,\mu))(\{(x,f(x)): x\in A\}) 
&=(\id\times f)_\# \mu (\{(x,f(x)):x\in A\})\nonumber\\
&=\mu(A)\nonumber \\
&>\nu(A) \nonumber \\
&=(\id\times g)_\# \nu(\{(x,g(x)): x\in A\}) \nonumber\\
&\ge (\id\times g)_\#\nu (\{x,f(x): x\in A, f(x)=g(x)\})\nonumber \\
&=(\id\times g)_\#\nu (\{(x,f(x)): x\in A\})\label{pf: graph of g} \\
&=T((g,\nu))(\{(x,f(x)):x\in A\})\nonumber 
\end{align}
where~\eqref{pf: graph of g} follows from the fact $(\id\times g)_\#\nu$ is supported on the graph of $g$. 
Thus, $T((f,\mu))\neq T((g,\nu))$. 

For the second case, there exists Borel set $B$, such that $f(x)\neq g(x),\forall x\in B$ and $\mu(B)>0$. 
Thus, we have 
\[ (\id\times f)_\#\mu(\{(x,f(x)):x\in B\}) = \mu(B)>0 \]
and
\begin{align*}
(\id\times f)_\#\nu(\{(x,f(x)):x\in B\})&=\nu\{x:f(x)=g(x),x\in B\}\nonumber\\
&=\mu\{x: f(x)=g(x),x\in B\}\nonumber\\
&=0.\nonumber
\end{align*}
Thus $T((f,\mu))\neq T((g,\nu))$. 
Therefore $T$ is a 1-1 mapping. 
\end{proof}

In the space $\mathcal{M}_+(\Omega\times \mathbb{R}^k)$, we can define the following OPT problem.
For $\beta\in(0,\infty)$, we define
$D_{\beta}: (\Omega\times \mathbb{R}^k)^2\to \mathbb{R}_+$ by
$$D^p_\beta((x,\tilde{x}),(y,\tilde{y}))=\frac{1}{\beta}\|x-y\|^p+\|\tilde{x}-\tilde{y}\|^p.$$
It is straightforward to show $D$ is metric and is equivalent to the $\ell^p$ metric in $\Omega\times \mathbb{R}^k$. Thus the following OPT problem defines a (p-th power of a) metric in $\mathcal{M}_+(\Omega\times \mathbb{R}^k)$ by \cite[Appendix C]{bai2022sliced} or \cite[Theorem 2.2]{heinemann2023kantorovich}, where  $\hat\mu,\hat\nu\in\mathcal{M}_+(\Omega\times\mathbb{R}^k)$:
\begin{align}
\text{OPT}_{D_{\beta}^p,\lambda}(\hat\mu,\hat\nu)=\inf_{\hat\gamma\in\Pi_\leq(\hat\mu,\hat\nu)}\int_{(\Omega\times\bbR^k)^2} D^p_{\beta}((x,\tilde{x}),(y,\tilde{y}))\, \dd \hat\gamma + \lambda(\|\hat\mu\|_{\TV}+\|\hat\nu\|_{\TV}-2\|\hat\gamma\|_{\TV}). \label{eq: OPT 2}
\end{align}
Similar to~\cite[Proposition 3.3]{garcia2016continuum}, we will show the OPT distance \eqref{eq: OPT 2} and the $\PTLp{p}$ distance \eqref{eq: PTLP} are equivalent.

\begin{proposition}\label{Pro: OPT and PTLP}
Choose $(f,\mu),(g,\nu)\in \mathcal{Q}^+_p(\Omega;\mathbb{R}^k)$, let $\hat\mu=(\id\times f)_\#\mu,\hat\nu=(\id\times g)_\# \nu$. Define $F: \Pi_\leq(\mu,\nu)\to \Pi_\leq (\hat\mu,\hat\nu)$ by 
$$\gamma\mapsto \hat\gamma=F(\gamma):=((\id\times f),(\id\times g))_\#\gamma.$$ 
Then, $F$ is bijection. Furthermore, let 
$\hat C(\hat\gamma;\hat\mu,\hat\nu,D_\beta^p,\lambda)$ and $C(\gamma;(f,\mu),(g,\nu),\beta,\lambda)$ denote the objective function in \eqref{eq: OPT 2} and \eqref{eq: PTLP} respectively,
we have
$$\hat C(\hat\gamma;\hat\mu,\hat\nu,D_{\beta}^p,\lambda)=C(\gamma; (f,\mu),(g,\nu),\beta,\lambda).$$
Therefore, 
$$\PTLp{p}_{\beta,\lambda}((f,\mu),(g,\nu))=\mathrm{OPT}_{D^p_{\beta},\lambda}(\hat\mu,\hat\nu),$$
\end{proposition}

\begin{proof}
First, it is straightforward that $F$ is well defined. 
We start by showing that $F$ is injective.
If $\gamma_1,\gamma_2\in\Pi_{\leq}(\mu,\nu)$ and $\gamma_1\neq\gamma_2$ then there exists $A,B$ such that $\gamma_1(A\times B)\neq \gamma_2(A\times B)$.
Without loss of generality assume $\gamma_1(A\times B)> \gamma_2(A\times B)$.
Let $\hat{\gamma}_1=F(\gamma_1)$ and $\hat{\gamma}_2=F(\gamma_2)$.
Then,
\begin{align*}
\hat{\gamma}_1(A\times\bbR^k \times B\times \bbR^k) & = ((\id \times f),(\id\times g))_{\#}\gamma_1(A\times \bbR^k \times B \times \bbR^k) \\
 & = \gamma_1\lp\lb (x,y)\,:\, (x,f(x),y,g(y))\in A\times \bbR^k\times B \times \bbR^k\rb\rp \\
 & = \gamma_1(A\times B) \\
 & > \gamma_2(A\times B) \\
 & = \gamma_2\lp\lb (x,y)\,:\, (x,f(x),y,g(y))\in A\times \bbR^k\times B \times \bbR^k\rb\rp \\
 & = ((\id \times f),(\id\times g))_{\#}\gamma_2(A\times \bbR^k \times B \times \bbR^k) \\
 & = \hat{\gamma}_2(A\times\bbR^k \times B\times \bbR^k).
\end{align*}
So $\hat{\gamma}_1\neq\hat{\gamma}_2$, hence $F$ is injective.

To show surjectivity, take any $\hat\gamma\in \hat\Pi_\leq(\hat\mu,\hat\nu)$.
Take any $\hat{A}, \hat{B}$ such that $0=(\hat{\mu}\times\hat{\nu})(\hat{A}\times\hat{B}) = \mu(\hat{A})\nu(\hat{B})$.
Then either $\hat{\mu}(\hat{A})=0$ or $\hat{\nu}(\hat{B})=0$.
In the first case $\hat{\gamma}(\hat{A}\times\hat{B})\leq \hat{\gamma}(A\times (\Omega\times \bbR^k)) \leq \hat{\mu}(\hat{A})=0$.
Similarly, in the second case $\hat{\gamma}(\hat{A}\times\hat{B})\leq \hat{\nu}(\hat{B})=0$ and so $\spt(\hat{\gamma})\subseteq \spt(\hat{\mu}\times\hat{\nu})$.
It follows in a similar way that $\spt(F(\gamma))\subset \spt(\hat{\mu}\times\hat{\nu})$ for any $\gamma\in\Pi_{\leq}(\mu,\nu)$.

We define $\gamma\in \mathcal{M}_+(\Omega^2)$ by 
$\gamma(A)=\hat\gamma(\{(x,f(x)),(y,g(y)): (x,y)\in A\})$.
We have $\gamma \in\Pi_\leq(\mu,\nu)$. 
Take $\hat{A},\hat{B}\subset \Omega\times \bbR^k$ and we compare $F(\gamma)(\hat{A}\times\hat{B})$ with $\hat{\gamma}(\hat{A}\times\hat{B})$.
By the previous argument we only need consider $\hat{A}\in \spt(\hat{\mu})$ and $\hat{B}\in\spt(\hat{\nu)})$.
In particular, we may assume that $\hat{A} = \{(x,f(x))\,:\,x\in A\}$ and $\hat{B} = \{(y,g(y))\,:\,y\in B\}$ for some $A,B\subseteq \Omega$.
Now
\[ F\gamma(\hat{A}\times\hat{B}) = \gamma\lp\lb (x,y)\,:\, ((x,f(x)),(y,g(y)))\in\hat{A}\times\hat{B}\rb\rp = \gamma(A\times B) = \hat{\gamma}(\hat{A}\times\hat{B}). \]
Thus $F\gamma = \hat{\gamma}$ and so $F$ is surjective.

For each $\gamma\in\Pi_\leq(\mu,\nu)$ we continue to let $\hat\gamma=F(\gamma)$ denote the corresponding measure in $\Pi_\leq(\hat\mu,\hat\nu)$.
We have 
\begin{align}
\hat{C}(\hat\gamma;\hat\mu,\hat\nu,D_\beta^p,\lambda)&=\int_{(\Omega\times\mathbb{R}^k)^2}D_\beta^p((x,\tilde{x}),(y,\tilde{y})) \, \dd\hat\gamma(x,\tilde{x},y,\tilde{y})+\lambda (\|\hat\mu\|_{\TV}+\|\hat\nu\|_{\TV}-2\|\hat\gamma\|_{\TV}) \nonumber \\
&=\int_{\Omega^2}\frac{1}{\beta}\|x-y\|^p+\|f(x)-g(y)\|^p\, \dd\gamma(x,y)+\lambda (\|\mu\|_{\TV}+\|\nu\|_{\TV}-2\|\gamma\|_{\TV})\nonumber\\
&=C(\gamma;(f,\mu),(g,\nu),\beta,\lambda)
\end{align}
Combining with the fact that $\gamma\mapsto \hat{\gamma}$ is a bijection, we have 
$$\mathrm{OPT}_{\lambda}(\hat\mu,\hat\nu):=\PTLp{p}_{\beta,\lambda}((f,\mu),(g,\nu))$$
which completes the proof.
\end{proof}

\begin{remark} Proposition \ref{pro: OPT and PTLP mapping} and \ref{Pro: OPT and PTLP} imply that $(\mathcal{Q}^+_p(\Omega),\PTLp{p}_{\lambda,\beta})$ is a metric space when $\beta,\lambda\in(0,\infty)$ and therefore we can conclude Theorem \ref{thm: PTLP is a metric}.
\end{remark}

We now prove existence of minimizers for $\PTLp{p}$ problem.


\begin{proof}[Proof of Theorem \ref{Thm: PTLP minimizer}]
From Proposition \ref{Pro: OPT and PTLP}, we have that the $\PTLp{p}$ problem \eqref{eq: PTLP} admits a solution if and only if the OPT problem \eqref{eq: OPT 2} admits a solution. By the relation between OPT and OT (see, for example, \cite[section 2]{caffarelli2010free}, or \cite[Appendix B]{bai2022sliced}), one can convert the OPT problem \eqref{eq: OPT 2} to a classical OT problem defined on $(\Omega\times \mathbb{R}^k)\cup\{\hat\infty\}$ where $\hat\infty$ is an isolated point. It is lower semi-continuous and bounded from below. Thus by the classical result in optimal transport theory (e.g.~\cite[Theorem 4.1]{Villani2009Optimal}), there exists an optimal transportation plan for the OPT problem \eqref{eq: OPT 2}). 
Note, an equivalent way to prove the existence of a minimizer is using the direct method from the calculus of variations (compactness and lower-semi continuity implies existence of minimizers). Indeed, $\Pi_\leq(\mu,\nu)$ is compact in the weak topology by Proposition \ref{pro: compact set} and $\gamma\to \hat{C}(\gamma;\hat{\mu},\hat{\nu},D_\beta^p,\lambda)$ is lower-semi continuous (in the sense of the weak topology). Thus the $\PTLp{p}$ problem \eqref{eq: PTLP} admits a minimizer. 

For the Empirical $\PTLp{p}$ problem, by the relation between $\PTLp{p}$ and OPT as discussed in Proposition \ref{Pro: OPT and PTLP} or Lemma \ref{lem: PTLP truncated} in the next section, it suffices to show there exists an 1-1 mapping that can solve the corresponding OPT problem (see \eqref{eq: OPT 2}). 
By \cite[Theorem 4.1]{bai2022sliced}, we complete the proof. 
\end{proof}

\section{The \texorpdfstring{$\PTLp{p}$}{PTLp} Problem for Extreme \texorpdfstring{$\beta$}{beta}}\label{sec: PTLP extreme beta}

In this section, we discuss the $\PTLp{p}$ problem when $\beta\to0$ and $\beta\to\infty$.
First, we prove equivalence of the $\PTLp{p}$ problem with a truncated version.

\begin{lemma}\label{lem: PTLP truncated} 
There exists optimal $\gamma\in \cM_+(\Omega^2)$ for the $\PTLp{p}$ problem \eqref{eq: PTLP} such that $\gamma(S)=0$ where $S:=\{(x,y): \frac{1}{\beta}\|x-y\|^p+\|f(x)-g(y)\|^p\ge 2\lambda\}$. 
Therefore, the $\PTLp{p}$ problem can be defined as 
\begin{align}
\PTLp{p}_{\beta,\lambda}((f,\mu),(g,\nu))&=\inf_{\gamma\in\Pi_\leq(\mu,\nu)}\int_{\Omega^2}\lp \frac{1}{\beta}\|x-y\|^p+\|f(x)-g(y)\|^p \rp \wedge 2\lambda \, \dd \gamma(x,y) \nonumber\\
&\quad+\lambda(\|\mu\|_{\TV}+\|\nu\|_{\TV}-2\|\gamma\|_{\TV}) \label{eq: PTLP 2}
\end{align}
\end{lemma}

\begin{proof}
Similar to the last section, the $\PTLp{p}$ problem can be written as the following OPT problem between $\mu,\nu$ defined as follows: 
\begin{align}
\text{OPT}_{c_{\beta,f,g},\lambda}(\mu,\nu):=\inf_{\gamma\in\Pi_{\leq}(\mu,\nu)}\int_{\Omega^2} c_{f,g}(x,y) \, \dd \gamma(x,y) + \lambda(\|\mu\|_{\TV}+\|\nu\|_{\TV}-2\|\gamma\|_{\TV}) \label{eq: OPT 3}
\end{align}
where the ground cost is defined as  $$c_{\beta,f,g}(x,y):=\frac{1}{\beta}\|x-y\|^p+\|f(x)-g(y)\|^p.$$
Thus, by~\cite[Lemma 3.2]{bai2022sliced}, we have that there exists an optimal $\gamma$ such that $\gamma(S)=0$ on $S=\{(x,y): c(x,y)\ge 2\lambda\}$ and we complete the proof. 
\end{proof}

Now we discuss the extreme cases $\beta=0$ and $\beta=\infty$. 


\begin{theorem}\label{thm: PTLP beta 0}
For any positive sequence $\beta_n\to 0$, we have 
\begin{equation}
    \lim_{n\to \infty}\PTLp{p}_{\beta,\lambda}((f,\mu),(g,\nu))=\|f-g\|_{\Lp{p}(\mu\wedge\nu),2\lambda}^p+\lambda\|\mu-\nu\|_{\TV} \label{eq: PTLP infty 1}
\end{equation}
\end{theorem}


\begin{proof} 
Let $\gamma=(\id\times\id)_{\#}(\mu\wedge \nu)$, then $\gamma\in\Pi_{\leq}(\mu,\nu)$ and
\begin{align*}
\PTLp{p}_{\beta_n,\lambda}((f,\mu),(g,\nu)) & \leq \int_{\Omega\times\Omega} \lp \frac{1}{\beta_n}\|x-y\|^p + \|f(x)-g(y)\|^p\rp \wedge 2\lambda \, \dd\gamma(x,y) \\
 & \qquad + \lambda\lp \|\mu\|_{\TV} + \|\nu\|_{\TV} - 2\|\gamma\|_{\TV} \rp \\
 & = \int_\Omega \|f(x)-g(x)\|^p\wedge 2\lambda \, \dd (\mu\wedge\nu)(x) + \lambda\lp \|\mu\|_{\TV}+\|\nu\|_{\TV}-2\|\mu\wedge\nu\|_{\TV}\rp \\
 & = \|f-g\|_{\Lp{p}(\mu\wedge\nu),2\lambda}^p + \lambda\|\mu-\nu\|_{\TV}.
\end{align*}
In the rest of the proof we will prove the coverse inequality.

Since
\begin{align*}
& \frac{1}{\beta_n} \lp \int_{\Omega^2} \|x-y\|^p\wedge 2\lambda \, \dd \gamma(x,y) + \beta_n\lambda \lp \|\mu\|_{\TV}+\|\nu\|_{\TV}-2\|\gamma\|_{\TV}\rp\rp \\
& \qquad \qquad \leq \int_{\Omega^2} \frac{1}{\beta_n} \|x-y\|^p\wedge 2\lambda + \|f(x)-g(y)\|^p\wedge 2\lambda \, \dd \gamma(x,y) + \lambda \lp \|\mu\|_{\TV}+\|\nu\|_{\TV}-2\|\gamma\|_{\TV}\rp
\end{align*} 
then
\[ \frac{1}{\beta_n}\mathrm{OPT}_{\|\cdot\|^p\wedge2\lambda,\lambda}(\mu,\nu)\leq \PTLp{p}_{\beta_n,\lambda}((f,\mu),(g,\nu)) \leq \|f-g\|_{\Lp{p}(\mu\wedge\nu),2\lambda}^p + \lambda\|\mu-\nu\|_{\TV}. \]
So $\mathrm{OPT}_{\|\cdot\|^p,\lambda}(\mu,\nu)=\mathrm{OPT}_{\|\cdot\|^p\wedge2\lambda,\lambda}(\mu,\nu) \leq O(\beta_n) \to 0$.

Let $\gamma_n\in\Pi_{\leq}(\mu,\nu)$ satisfy
\[ \mathrm{OPT}_{\|\cdot\|^p,\beta_n\lambda}^p(\mu,\nu) = \int_{\Omega^2} \|x-y\|^p\wedge 2\lambda \, \dd \gamma_n(x,y) + \beta_n\lambda \lp \|\mu\|_{\TV} + \|\nu\|_{\TV} - 2\|\gamma_n\|_{\TV}\rp. \]
As $\Pi_{\leq}(\mu,\nu)$ is weakly sequentially compact then there exists a subsequence (which we relabel) and a $\gamma_0$ such that $\gamma_n\rightharpoonup\gamma_0\in\Pi_{\leq}(\mu,\nu)$.
Now, $\|\gamma_n\|_{\TV}=\gamma_n(\Omega^2)\to \gamma_0(\Omega^2) = \|\gamma_0\|_{\TV}$, and $(x,y)\mapsto \|x-y\|^p\wedge2\lambda$ is continuous and bounded, so
\[ \mathrm{OPT}_{\|\cdot\|^p\wedge 2\lambda,\beta_n\lambda}(\mu,\nu) \to \int_{\Omega^2} \|x-y\|^p\wedge 2\lambda \, \dd \gamma_0(x,y) \]
Hence, $\int_{\Omega^2} \|x-y\|^p \, \dd \gamma_0(x,y)=0$ and so $x=y$ $\gamma_0$-a.e..
In particular, there exists $\mu_0$ such that $\gamma_0=(\id\times\id)_\#\mu_0$ and $\mu_0\leq \nu$, $\mu_0\leq \nu$.
We are left to show $\lim_{n\to\infty} \PTLp{p}_{\beta_n,\lambda}((f,\mu),(g,\nu)) \geq \PTLp{p}_{0,\lambda}((f,\mu),(g,\nu))$, where
\begin{align}
\PTLp{p}_{0,\lambda}((f,\mu),(g,\nu)):=\inf_{\mu_0\leq \mu\wedge\nu}\int_{\Omega}\|f(x)-g(x)\|^p\wedge 2\lambda \, \dd \mu_0(x) + \lambda(\|\mu\|_{\TV}+\|\nu\|_{\TV}-2\|\mu_0\|_{\TV}).  \label{pf: PTLp beta0 orig}
\end{align}
We will show that the minimizer is $\mu\wedge \nu$. Indeed, let $C'(\mu_0;(f,\mu),(g,\nu),0,\lambda)$ denote the transportation cost induced by $\mu_0$, we have 
\begin{align}
 &C'(\mu\wedge \nu;(f,\mu),(g,\nu),0,\lambda)-C'(\mu_0;(f,\mu),(g,\nu),0,\lambda) \nonumber \\
 & \qquad =\int_{\Omega}\|f(x)-g(x)\|^p\wedge 2\lambda -2\lambda \, \dd (\mu\wedge \nu -\mu_0)(x) \nonumber\\
 & \qquad \leq 0 \nonumber
\end{align}
where the inequality follows since $\mu\wedge \nu -\mu_0$ is a nonnegative measure. 
Thus we have 
\begin{align}
  PTL^p_{0,\lambda}((f,\mu),(g,\nu))=\|f-g\|_{\mu\wedge \nu,2\lambda}+\lambda(\|\mu-\nu\|_{TV}).\label{pf: PTLp beta0} 
\end{align}

Since $\Ckb{0}(\Omega)$ is dense in $\Lp{p}(\mu),\Lp{p}(\nu)$, for any $\eps>0$ there exists $f_\eps,g_\eps \in \Ckb{0}(\Omega)$ such that $$\int_{\Omega^2}\|f(x)-f_\eps(x)\|^p\, \dd \mu(x) \leq \eps, \qquad \int_{\Omega^2}\|g(y)-g_\eps(y)\|^p \, \dd \nu(y) \leq\eps.$$

Now,
\begin{align*}
\lp \int_{\Omega^2} \|f(x)-g(y)\|^p \, \dd\gamma_n(x,y) \rp^{\frac{1}{p}} & \geq \lp \int_{\Omega^2} \|f_\eps(x)-g_\eps(y)\|^p \, \dd\gamma_n(x,y) \rp^{\frac{1}{p}} \\
 & \qquad - \lp \int_{\Omega^2} \|f(x)-f_\eps(x)\|^p \, \dd\gamma_n(x,y) \rp^{\frac{1}{p}} \\
 & \qquad - \lp \int_{\Omega^2} \|g_\eps(y)-g(y)\|^p \, \dd\gamma_n(x,y) \rp^{\frac{1}{p}} \\
 & \geq \lp \int_{\Omega^2} \|f_\eps(x)-g_\eps(y)\|^p \, \dd\gamma_n(x,y) \rp^{\frac{1}{p}} - 2\eps \\
 & \to \lp \int_{\Omega^2} \|f_\eps(x)-g_\eps(y)\|^p \, \dd\gamma_0(x,y) \rp^{\frac{1}{p}} - 2\eps.
\end{align*}
So, 
\begin{align*}
\liminf_{n\to\infty} \PTLp{p}_{\beta_n,\lambda}((f,\mu),(g,\nu)) & \geq \int_{\Omega^2} \|f_\eps(x)-g_\eps(y)\|^p \, \dd\gamma_0(x,y) - K\eps \\
 & \qquad \qquad + \lambda\lp \|\mu\|_{\TV}+\|\nu\|_{\TV}-2\|\gamma_0\|_{\TV}\rp \\
 & \geq \int_{\Omega^2} \|f(x)-g(y)\|^p \, \dd\gamma_0(x,y) - K^\prime \eps \\
 & \qquad \qquad + \lambda\lp \|\mu\|_{\TV}+\|\nu\|_{\TV}-2\|\gamma_0\|_{\TV}\rp \\
 & \geq \PTLp{p}_{0,\lambda}((f,\mu),(g,\nu)) - K^\prime\eps
\end{align*}
for constants $K,K^\prime$.
Taking $\eps\to 0$ completes the proof.
\end{proof}
Based on the above theorem, we can extend the $\PTLp{p}$ distance for $\beta=0$.


\begin{corollary}
When $\mu=\nu$, we have 
$$\PTLp{p}_{0,\lambda}((f,\mu),(g,\mu))=\|f-g\|_{\Lp{p}(\mu),2\lambda}^p.$$ 
%
\end{corollary}
The proof is straightforward. 

Next, we discuss the case $\beta=\infty$.

\begin{theorem}\label{thm: PTLP beta infty}
For any positive sequence  $\beta_n\to \infty$, we have 
\begin{align}
\lim_{n\to\infty}\PTLp{p}_{\beta,\lambda_n}((f,\mu),(g,\nu))=\mathrm{OPT}_{\lambda}(f_\#\mu,g_\#\nu). \label{eq: PTLP beta infty obj}
\end{align}
\end{theorem}

\begin{proof}
Without loss of generality, we can assume $\beta_n$ is a monotonic increasing sequence. 
Note, it is straightforward to show (e.g. see Proposition \ref{Pro: OPT and PTLP} by setting the $\frac{1}{\beta}$ term to be $0$):
\begin{align}
  \text{OPT}_{\lambda}(f_\#\mu,g_\#\nu)&=\text{OPT}_{c_{\infty,f,g},\lambda}(\mu,\nu)\nonumber \\ 
  &=\inf_{\gamma\in\Pi_{\leq}(\mu,\nu)} \int_{\Omega^2}\|f(x)-g(y)\|^p \, \dd\gamma(x,y)+ \lambda\lp\|\mu\|_{\TV}+\|\nu\|_{\TV}-2\|\gamma\|_{\TV}\rp  \label{eq: OPT 4}
\end{align}
where $c_{\infty,f,g}(x,y):=\|f(x)-g(y)\|^p$. 
For each $n$ and $\gamma$, we have  
\begin{align}
&\int_{\Omega^2}\frac{1}{\beta_n} \|x-y\|^p+\|f(x)-g(y)\|^p \, \dd \gamma(x,y) +\lambda(\|\mu\|_{\TV}+\|\nu\|_{\TV}-2\|\gamma\|_{\TV}) \nonumber\\ 
&\ge \int_{\Omega^2}\|f(x)-g(y)\|^p \, \dd \gamma(x,y) +\lambda(\|\mu\|_{\TV}+\|\nu\|_{\TV}-2\|\gamma\|_{\TV}).
\end{align}
Taking the infimum on both sides and passing to the limit, we have 
\begin{align}
\liminf_{n\to\infty}\PTLp{p}_{\beta,\lambda_n}((f,\mu),(g,\nu))\ge\mathrm{OPT}_{c_{f,g},\lambda}(\mu,\nu) \label{pf: PTlp beta infty lowerbound}     
\end{align}
For the other direction, we let $\gamma_\infty$ be an optimal transportation plan for $\mathrm{OPT}_{c_{f,g},\lambda}(\mu,\nu)$, then 
\begin{align}
&\limsup_{n\to\infty}\PTLp{p}_{\beta,\lambda_n}((\mu,f),(\nu,g))\nonumber \\ 
&\leq \limsup_{n\to\infty} \int_{\Omega^2}\lp\frac{1}{\beta_n}\|x-y\|^p+\|f(x)-g(y)\|^p\rp\wedge 2\lambda \, \dd \gamma_\infty(x,y)+\lambda(\|\mu\|_{\TV}+\|\nu\|_{\TV}-2\|\gamma_\infty\|_{\TV})\nonumber  \\ 
&= \int_{\Omega^2}\limsup_{n\to\infty} \lp\frac{1}{\beta_n}\|x-y\|^p+\|f(x)-g(y)\|^p\rp \wedge 2\lambda \, \dd\gamma_\infty(x,y)+\lambda(\|\mu\|_{\TV}+\|\nu\|_{\TV}-2\|\gamma_\infty\|_{\TV}) \nonumber  \\ 
&=\int_{\Omega^2}\|f(x)-g(y)\|\wedge 2\lambda \, \dd \gamma_\infty(x,y)+\lambda(\|\mu\|_{\TV}+\|\nu\|_{\TV}-2\|\gamma_\infty\|_{\TV})\nonumber
\end{align}
where the third line follows from the Monotone convergence theorem (or Beppo Levi's lemma). Thus 
\begin{align}
\limsup_{n\to\infty}\PTLp{p}_{\beta,\lambda_n}((f,\mu),(g,\nu))\leq  \mathrm{OPT}_{c_{\infty,f,g},\lambda}(\mu,\nu) \label{pf: PTLp beta infty upper bound}
\end{align}
Combining~\eqref{pf: PTlp beta infty lowerbound} and~\eqref{pf: PTLp beta infty upper bound}, we complete the proof. 
\end{proof}

\begin{remark}
Combining Theorem \ref{thm: PTLP beta 0} and \ref{thm: PTLP beta infty}, we prove Theorem \ref{thm: PTLP extreme beta} in the main text.    
\end{remark}

\newpage
\section{More 1 NN Classification Experiments}
We provide additional nearest neighbor classification results in Table \ref{tab:2}.

\begin{table}[H]
    \label{tab:2}
    \centering
    \caption{Additional nearest neighbor classification results on the modified UCR dataset \cite{tan2019time}. Similarly, we highlight top two performers and provide the averages.}
    \small
    \scalebox{0.82}{
    \begin{tabular}{cc|c|c|c|c|c|c|c|c|c|c|c}
        \hline
        \multirow{18}{*}{{\rotatebox[origin=c]{90}{\textbf{Prefix Dataset}}}} & Method & $\PTLp{p}$& $\SPTLp$ & $\TLp$ & $\STLp$ & $\OT$&  \thead{SDTW\\{\tiny $\gamma=0.01$}}  &  \thead{SDTW\\{\tiny $\gamma=1$}} & DTW &WDTW & DDTW & $\Lp{p}$  \\ \hline
        &Coffee & \textbf{0.79} & 0.68 & 0.50 & 0.61 & 0.54 & 0.64 & 0.57 & 0.64 & 0.54 & \textbf{0.71} & 0.61 \\ 
        &ECGFiveDays & 0.58 & 0.60 & \textbf{0.63} & \textbf{0.63} & 0.51 & 0.61 & 0.58 & 0.61 & 0.50 & 0.59 & 0.55 \\ 
        &DistalPhalanxTW & 0.53 & \textbf{0.54} & 0.50 & 0.51 & 0.45 & 0.51 & 0.49 & \textbf{0.55} & 0.30 & 0.30 & 0.53 \\ 
        &BirdChicken & \textbf{0.75} & \textbf{0.75} & 0.55 & 0.55 & 0.60 & 0.70 & 0.70 & 0.70 & 0.50 & 0.65 & 0.55 \\ 
        &GunPoint & 0.63 & \textbf{0.71} & 0.65 & 0.66 & 0.67 & 0.65 & 0.69 & \textbf{0.72} & 0.49 & 0.49 & 0.65 \\ 
        &CBF & 0.81 & 0.81 & 0.71 & 0.72 & 0.54 & \textbf{0.86} & 0.84 & \textbf{0.86} & 0.33 & 0.66 & 0.46 \\ 
        &DPOC & 0.62 & \textbf{0.72} & 0.64 & 0.64 & 0.59 & 0.66 & \textbf{0.67} & \textbf{0.67} & 0.58 & 0.58 & 0.57 \\ 
        &OliveOil & \textbf{0.33} & 0.30 & \textbf{0.33} & \textbf{0.33} & 0.30 & 0.27 & 0.23 & 0.23 & 0.17 & 0.17 & 0.13 \\ 
        &Symbols & \textbf{0.64} & 0.61 & 0.42 & 0.56 & 0.32 & \textbf{0.64} & \textbf{0.64} & 0.62 & 0.16 & 0.56 & 0.47 \\ 
        &FaceFour & \textbf{0.66} & 0.53 & 0.56 & \textbf{0.59} & 0.25 & 0.48 & 0.59 & 0.49 & 0.30 & 0.26 & 0.27 \\ 
        &ItalyPowerDemand & \textbf{0.79} & \textbf{0.81} & 0.65 & 0.68 & 0.62 & 0.71 & 0.74 & 0.71 & 0.50 & 0.69 & 0.61 \\ 
        &FISH & \textbf{0.55} & \textbf{0.53} & 0.38 & 0.38 & 0.26 & 0.47 & 0.52 & 0.44 & 0.14 & 0.49 & 0.39 \\ 
        &SwedishLeaf & \textbf{0.64} & 0.62 & 0.47 & 0.48 & 0.31 & 0.45 & 0.51 & 0.43 & 0.06 & \textbf{0.66} & 0.24 \\ 
        &ToeSegmentation1 & 0.58 & 0.65 & \textbf{0.76} & \textbf{0.72} & 0.48 & 0.59 & 0.68 & 0.58 & 0.53 & 0.65 & 0.48 \\ 
        &ToeSegmentation2 & 0.62 & 0.75 & 0.62 & 0.59 & 0.54 & 0.64 & 0.75 & 0.64 & \textbf{0.82} & \textbf{0.82} & 0.78 \\ \hline
        &\thead{Average\\ Prefix}  & \textbf{0.64} & \textbf{0.64} & 0.56 & 0.58 & 0.47 & 0.59 & 0.61 & 0.59 & 0.39 & 0.55 & 0.49 \\ 
        \hline 
        \hline
        \multirow{15}{*}{{\rotatebox[origin=c]{90}{\textbf{Subsequence Dataset}}}}
        &Coffee & 0.54 & 0.57 & 0.54 & \textbf{0.71} & 0.46 & \textbf{0.71} & 0.57 & \textbf{0.71} & 0.54 & 0.50 & 0.64 \\ 
        &ECGFiveDays & 0.57 & 0.56 & 0.57 & 0.57 & 0.52 & \textbf{0.59} & 0.57 & 0.58 & 0.50 & \textbf{0.58} & 0.53 \\ 
        &DistalPhalanxTW & 0.51 & 0.56 & 0.55 & \textbf{0.58} & 0.40 & \textbf{0.61} & 0.45 & 0.48 & 0.30 & 0.51 & 0.31 \\ 
        &BirdChicken & 0.40 & 0.55 & 0.50 & 0.60 & 0.35 & \textbf{0.70} & \textbf{0.75} & \textbf{0.70} & 0.50 & 0.50 & 0.60 \\ 
        &GunPoint & 0.65 & 0.66 & 0.63 & 0.64 & 0.56 & 0.65 & \textbf{0.71} & 0.67 & 0.49 & \textbf{0.77} & 0.59 \\ 
        &CBF & 0.71 & 0.71 & 0.65 & 0.70 & 0.48 & \textbf{0.75} & 0.74 & \textbf{0.75} & 0.33 & 0.55 & 0.36 \\ 
        &DPOC & 0.60 & \textbf{0.63} & 0.62 & 0.62 & \textbf{0.62} & 0.62 & 0.62 & 0.59 & 0.58 & 0.58 & 0.60 \\ 
        &OliveOil & \textbf{0.50} & 0.43 & \textbf{0.53} & 0.40 & 0.43 & 0.50 & 0.50 & 0.37 & 0.17 & 0.17 & 0.37 \\ 
        &Symbols & \textbf{0.52} & 0.50 & 0.49 & 0.48 & 0.37 & 0.48 & 0.51 & 0.48 & 0.16 & \textbf{0.68} & 0.30 \\ 
        &FaceFour & \textbf{0.51} & 0.49 & \textbf{0.52} & 0.40 & 0.39 & 0.35 & 0.38 & 0.35 & 0.30 & 0.40 & 0.26 \\ 
        &ItalyPowerDemand & 0.66 & \textbf{0.71} & 0.61 & 0.63 & 0.60 & 0.70 & \textbf{0.71} & 0.70 & 0.50 & 0.68 & 0.52 \\ 
        &FISH & 0.35 & \textbf{0.40} & 0.31 & 0.31 & 0.17 & 0.31 & 0.30 & 0.27 & 0.14 & \textbf{0.57} & 0.21 \\ 
        &SwedishLeaf & 0.42 & \textbf{0.44} & 0.39 & 0.40 & 0.20 & 0.33 & 0.38 & 0.31 & 0.06 & \textbf{0.60} & 0.11 \\ 
        &ToeSegmentation1 & 0.67 & 0.68 & 0.66 & 0.67 & 0.55 & 0.67 & \textbf{0.69} & 0.66 & 0.53 & \textbf{0.69} & 0.48 \\ 
        &ToeSegmentation2 & 0.71 & 0.78 & 0.70 & 0.68 & 0.58 & 0.69 & 0.79 & 0.69 & \textbf{0.82} & \textbf{0.82} & 0.73 \\ 
        \hline
      &  \thead{Average\\Subsequence} & 0.55 & \textbf{0.58} & 0.55 & 0.56 & 0.45 & 0.58 & \textbf{0.58} & 0.56 & 0.39 & 0.57 & 0.44 \\ 
      \hline
      \hline
      \multirow{15}{*}{{\rotatebox[origin=c]{90}{\textbf{Suffix Dataset}}}} 
        &Coffee & 0.54 & 0.57 & 0.61 & 0.61 & 0.54 & \textbf{0.68} & \textbf{0.68} & \textbf{0.71} & 0.54 & 0.66 & 0.57 \\ 
        &ECGFiveDays & 0.57 & 0.61 & 0.55 & 0.58 & 0.52 & \textbf{0.63} & 0.62 & \textbf{0.62} & 0.50 & 0.50 & 0.53 \\ 
        &DistalPhalanxTW & 0.56 & 0.58 & 0.53 & 0.55 & 0.50 & \textbf{0.61} & \textbf{0.58} & 0.57 & 0.30 & 0.30 & 0.28 \\ 
        &BirdChicken & 0.45 & 0.55 & \textbf{0.70} & \textbf{0.60} & 0.40 & 0.45 & 0.45 & 0.45 & 0.50 & 0.50 & 0.50 \\ 
        &GunPoint & 0.61 & \textbf{0.65} & 0.57 & 0.60 & 0.65 & 0.65 & \textbf{0.66} & 0.62 & 0.49 & 0.49 & 0.58 \\ 
        &CBF & 0.61 & 0.72 & 0.71 & 0.73 & 0.42 & \textbf{0.78} & 0.78 & \textbf{0.78} & 0.33 & 0.33 & 0.34 \\ 
        &DPOC & 0.68 & \textbf{0.71} & 0.67 & 0.68 & 0.59 & 0.70 & \textbf{0.71} & 0.68 & 0.58 & 0.58 & 0.55 \\ 
        &OliveOil & 0.17 & \textbf{0.30} & 0.23 & \textbf{0.30} & 0.27 & 0.17 & 0.20 & 0.13 & 0.17 & 0.13 & 0.17 \\ 
        &Symbols & \textbf{0.55} & 0.52 & 0.51 & 0.53 & 0.38 & 0.51 & \textbf{0.55} & 0.50 & 0.16 & 0.53 & 0.24 \\ 
        &FaceFour & 0.38 & \textbf{0.45} & 0.27 & 0.28 & 0.24 & 0.35 & \textbf{0.39} & 0.35 & 0.30 & \textbf{0.39} & 0.17 \\ 
        &ItalyPowerDemand & 0.80 & 0.81 & \textbf{0.82} & 0.77 & 0.69 & 0.82 & \textbf{0.85} & 0.82 & 0.50 & 0.77 & 0.59 \\ 
        &FISH & 0.36 & 0.37 & 0.35 & 0.35 & 0.16 & 0.37 & \textbf{0.38} & 0.34 & 0.14 & \textbf{0.45} & 0.14 \\ 
        &SwedishLeaf & 0.53 & 0.54 & 0.53 & 0.53 & 0.30 & 0.49 & \textbf{0.57} & 0.48 & 0.06 & \textbf{0.68} & 0.18 \\ 
        &ToeSegmentation1 & 0.59 & \textbf{0.64} & 0.60 & 0.51 & 0.52 & 0.60 & 0.62 & 0.60 & 0.53 & \textbf{0.82} & 0.52 \\ 
        &ToeSegmentation2 & 0.71 & \textbf{0.77} & 0.62 & 0.62 & 0.55 & 0.68 & 0.73 & 0.69 & \textbf{0.82} & 0.69 & \textbf{0.77} \\ 
         \hline
     &   \thead{Average\\ Suffix} & 0.54 & \textbf{0.59} & 0.55 & 0.55 & 0.45 & 0.57 & \textbf{0.58} & 0.56 & 0.39 & 0.52 & 0.41 \\  \hline
      & \thead{Average\\ Total} & 0.58 & \textbf{0.60} & 0.55 & 0.56 & 0.46 & 0.58 & \textbf{0.59} & 0.57 & 0.39 & 0.55 & 0.45
    \\ \hline         
    \end{tabular}}
\end{table}

\end{document}